\documentclass[letterpaper]{article} 
\usepackage[submission]{aaai24}  
\usepackage{times}  
\usepackage{helvet}  
\usepackage{courier}  
\usepackage[hyphens]{url}  
\usepackage{graphicx} 
\usepackage{natbib}  
\usepackage{caption} 
\usepackage{algorithm}
\usepackage{algorithmic}

\usepackage{newfloat}
\usepackage{listings}
\DeclareCaptionStyle{ruled}{labelfont=normalfont,labelsep=colon,strut=off} 
\floatstyle{ruled}
\newfloat{listing}{tb}{lst}{}
\floatname{listing}{Listing}
\title{Learning MDL Logic Programs From Noisy Data}

\author{
    C\'{e}line Hocquette\textsuperscript{\rm 1}, Andreas Niskanen\textsuperscript{\rm 2},
    Matti J\"{a}rvisalo\textsuperscript{\rm 2},
    Andrew Cropper\textsuperscript{\rm 1}
}

\affiliations {
    \textsuperscript{\rm 1}University of Oxford\\
    \textsuperscript{\rm 2}University of Helsinki\\
    celine.hocquette@cs.ox.ac.uk, andreas.niskanen@helsinki.fi,
    matti.jarvisalo@helsinki.fi,
    andrew.cropper@cs.ox.ac.uk
}

\usepackage[utf8]{inputenc}

\usepackage{xcolor}
\usepackage{booktabs}
\usepackage{inconsolata}
\usepackage{tikz}
\usepackage{pgfplots}
\usepackage{amsmath}
\usepackage{microtype}
\usepackage{cancel}
\usepackage{amssymb}
\usepackage[titletoc]{appendix}

\newcommand{\name}{\textsc{MaxSynth}}

\newcommand{\popper}{\textsc{Popper}}

\newcommand{\noisypopper}{\textsc{NoisyPopper}}

\newcommand{\ilasp}{\textsc{Ilasp}}
\newcommand{\ale}{\textsc{Aleph}}

\newcommand{\deltailp}{\textsc{$\delta$ILP}}
\newcommand{\aspal}{\textsc{Aspal}}
\newcommand{\combo}{\textsc{Combo}}

\usepackage{amsthm}
\theoremstyle{definition}
\newtheorem{definition}{Definition}

\newtheorem{theorem}{Theorem}
\newtheorem{proposition}{Proposition}
\newtheorem{lemma}{Lemma}
\newtheorem{assumption}{Assumption}

\usepackage{pgfkeys}
    \newenvironment{customlegend}[1][]{%
        \begingroup
        \csname pgfplots@init@cleared@structures\endcsname
        \pgfplotsset{#1}%
    }{%
        \csname pgfplots@createlegend\endcsname
        \endgroup
    }%
    \def\addlegendimage{\scriptsize\csname pgfplots@addlegendimage\endcsname}
\usepackage{comment}

\usepackage{listings}
\lstnewenvironment{myalgorithm}[1][] 
{
    \lstset{ 
        basicstyle=\ttfamily\small,
        columns=flexible,
        showspaces=false,               
        showstringspaces=false,         
        mathescape=true,
        numbers=left,
        escapeinside={*}{*},
        columns=flexible,
        escapechar=@,
        numbersep=3pt,        
        keywordstyle=\bfseries,
        keywords={,and, return, not, def, in,elif,if, else, for, foreach, while,}
        numbers=left,
        xleftmargin=8pt,
        #1 
    }
}
{}

\usepackage{bibentry}

\begin{document}

\maketitle

\begin{abstract}
Many inductive logic programming approaches struggle to learn programs from noisy data.
To overcome this limitation, we introduce an approach that learns minimal description length programs from noisy data, including recursive programs.
Our experiments on several domains, including drug design, game playing, and program synthesis, show that our approach can outperform existing approaches in terms of predictive accuracies and scale to moderate amounts of noise.
 \end{abstract}
\section{Introduction}

The goal of inductive logic programming (ILP) \cite{mugg:ilp} is to induce a logic program (a set of logical rules) that generalises training examples and background knowledge.
A common criticism of ILP is that it cannot handle noisy data \cite{dilp,DBLP:conf/kr/CucalaGM22}.
This criticism is unfounded: most ILP approaches can learn from noisy data \cite{ilpintro}.
For instance, set-covering approaches \cite{progol,aleph,atom,quickfoil,probfoil} search for rules that generalise a subset of the examples.

Although most ILP approaches can learn from noisy data, they struggle to learn recursive programs and perform predicate invention, two important features when learning complex algorithms \cite{metabias,metaopt}. 
Moreover, they are not guaranteed to learn optimal programs, such as textually minimal programs, and tend to overfit.

Recent approaches overcome these limitations and can learn recursive and textually minimal programs \cite{aspal,hexmil} and perform predicate invention \cite{mugg:metagold,hopper}.
However, these approaches struggle to learn from noisy data because they search for a program that strictly generalises all the positive and none of the negative examples.

In this paper, our goal is to learn recursive programs and support predicate invention in a noisy setting.
Following \citet{combo}, we first search for small programs that generalise a subset of the examples.
We then search for a combination of these smaller programs to form a larger program.
\citet{combo} search for a combination that strictly generalises all the positive and none of the negative examples, i.e. they cannot learn from noisy data.
By contrast, we relax this condition to learn from noisy data.
To avoid overfitting, we search for a combination that trades off model complexity (program size) and data fit (training accuracy).
To do so, we use the minimal description length (MDL) principle \cite{mdl}.
In other words, we introduce an approach that learns MDL programs from noisy data.

To explore our idea, we build on \emph{learning from failures} (LFF) \cite{popper}.
LFF frames the ILP problem as a constraint satisfaction problem (CSP), where each solution to the CSP represents a program (a hypothesis).
The goal of a LFF learner is to accumulate constraints to restrict the hypothesis space (the set of all hypotheses) and thus constrain the search.
We use LFF to explore our idea because it can learn recursive programs and perform predicate invention.
We build on LFF by learning MDL programs from noisy examples.
We introduce constraints which are optimally sound in that they do not prune MDL programs.
To find an MDL combination, we use a maximum satisfiability (MaxSAT) solver~\cite{maxsat}.

\subsubsection*{Novelty and contributions}
The main novelty of this paper is the idea of learning small programs from noisy examples and using a MaxSAT solver to find an MDL combination.
The benefits, which we show on diverse domains, are (i)~the ability to learn complex programs from noisy examples, and (ii)~improved performance compared to existing approaches.

Overall, our contributions are:
\begin{enumerate}
\item We introduce \name{}, which learns MDL programs from noisy examples, including recursive programs.
\item We introduce constraints for this noisy setting and prove that they are optimally sound (Propositions \ref{specialisation} and \ref{generalisation}).
\item We prove the correctness of \name{}, i.e. that it always learns an MDL program (Theorem \ref{correctness}).
\item We experimentally show on multiple domains, including drug design, game playing, and program synthesis, that \name{} can (i) substantially improve predictive accuracies compared to other systems, and (ii) scale to moderate amounts of noise (30\%).
We also show that our noisy constraints can reduce learning times by 99\%.
\end{enumerate}

\section{Related Work}

\textbf{ILP.}
Most ILP approaches support noise \cite{foil,progol,mccreath1997ilp,tilde,aleph,hypern,atom,quickfoil,probfoil}.
However, these approaches do not support predicate invention, struggle to learn recursive programs, and are not guaranteed to learn an MDL program.
Recent approaches can learn textually minimal and recursive programs but are not robust to noisy examples \cite{aspal,mugg:metagold,hexmil,popper,meta_abduce,hopper}.
There are two notable exceptions.
\deltailp{} \cite{dilp} frames the ILP problem as a differentiable neural architecture and is robust to noisy data.
\noisypopper{} \cite{noisypopper} can learn MDL and recursive programs from noisy examples.
However, these approaches can only learn programs with a small number of small rules.
For instance, \deltailp{} cannot learn programs with more than a few rules and can only use binary relations.
By contrast, \name{} can learn MDL programs with many rules and any arity relation.

\textbf{Rule selection.}
Many systems formulate the ILP problem as a rule selection problem \cite{aspal,hexmil,difflog,prosynth,apperception,shitruleselection}.
These approaches precompute every possible rule in the hypothesis space and then search (often using a constraint solver) for a subset that entails all the positive and none of the negative examples.
Some approaches relax this requirement to find a subset with the best coverage using solver optimisation \cite{noisyilasp,apperception} or numerical methods \cite{difflog}.
However, because they precompute every possible rule, these approaches cannot learn rules with a large number of literals.
By contrast, we do not precompute all possible rules.

\textbf{Sampling.}
Sampling can mitigate noise.
\citet{noisyprogramsynthesis} pair a data sampler which selects representative subsets of the data with a regularised program generator to avoid overfitting.
\textsc{Metagol$_{nt}$} \cite{mugg:vision} finds hypotheses consistent with randomly sampled subsets of the training examples and evaluates each resulting program on the remaining training examples.
\textsc{Metagol$_{nt}$} needs as input a parameter about the noise level.
By contrast, \name{} does not need a user-provided noise level parameter and is guaranteed to learn an MDL program.

\textbf{Rule mining.}
AMIE+ \cite{amie+} learns rules from noisy knowledge bases.
 However, AMIE+ can only use unary and binary relations, so it cannot be used on most of the datasets in our experiments, which require relations of arity greater than two.
By contrast, \name{} can learn programs with relations of any arity.

\textbf{MDL.}
Several approaches use cost functions based on MDL \cite{foil,progol,aleph,DBLP:conf/ijcai/HuangP07}.
However, they are not guaranteed to find a program that minimises this cost function because they greedily learn a single rule at a time.
By contrast, \name{} learns a global MDL program.
\citet{luc:mdl} learn propositional CNF using MDL.
By contrast, we learn first-order theories.


\section{Problem Setting}
We describe our problem setting.
We assume familiarity with logic programming \cite{lloyd:book} but have included a summary in the appendix.

\subsection{Learning From Failures}
We use the LFF setting.
A \emph{hypothesis} is a definite program with the least Herbrand model semantics.
A \emph{hypothesis space} $\mathcal{H}$ is a set of hypotheses.
LFF uses \emph{hypothesis constraints} to restrict the hypothesis space.
Let $\mathcal{L}$ be a meta-language that defines hypotheses.
For instance, consider a language with two literals \emph{h\_lit/3} and \emph{b\_lit/3} which represent \emph{head} and \emph{body} literals respectively.
With this language, we denote the rule \emph{last(A,B) $\leftarrow$ tail(A,C), head(C,B)} as the set of literals \emph{\{h\_lit(0,last,(0,1)), b\_lit(0,tail,(0,2)), b\_lit(0,head,(2,1))\}}.
The first argument of each literal is the rule index, the second is the predicate symbol, and the third is the literal variables, where \emph{0} represents \emph{A}, \emph{1} represents \emph{B}, etc.
A \emph{hypothesis constraint} is a constraint (a headless rule) expressed in $\mathcal{L}$.
Let $C$ be a set of hypothesis constraints written in a language $\mathcal{L}$.
A hypothesis is \emph{consistent} with $C$ if when written in $\mathcal{L}$ it does not violate any constraint in $C$.
We denote as $\mathcal{H}_{C}$ the subset of the hypothesis space $\mathcal{H}$ which does not violate any constraint in $C$.

We define a LFF input:

\begin{definition}[\textbf{LFF input}]
\label{def:probin}
A \emph{LFF input} is a tuple $(E, B, \mathcal{H}, C, cost)$ where $E=(E^{+}, E^{-})$ is a pair of sets of ground atoms denoting positive ($E^+$) and negative ($E^-$) examples, $B$ is a definite program denoting background knowledge,
$\mathcal{H}$ is a hypothesis space, $C$ is a set of hypothesis constraints,
and $cost$ is a function that measures the cost of a hypothesis.
\end{definition}

\noindent
We define a solution to a LFF input in the non-noisy setting:
\begin{definition}[\textbf{Non-noisy solution}]
\label{def:solution}
Given a LFF input $(E, B, \mathcal{H}, C, cost)$, where $E=(E^{+}, E^{-})$, a hypothesis $h \in \mathcal{H}_{C}$ is a \emph{non-noisy solution} when $h$ is \emph{complete} ($\forall e \in E^+, \; B \cup h \models e$) and \emph{consistent} ($\forall e \in E^-, \; B \cup h \not\models e$).
\end{definition}
\noindent
A hypothesis that is not a non-noisy solution is a \emph{failure}.
A LFF learner builds constraints from failures to restrict the hypothesis space.
For instance, if a hypothesis $h$ is inconsistent (entails a negative example), a generalisation constraint prunes generalisations of $h$ as they are also inconsistent.

In the non-noisy setting, a cost function only takes as input a hypothesis, i.e. they are of the type $cost : \mathcal{H} \mapsto \mathbb{N}$.
For instance, the cost of a hypothesis is typically measured as its size (the number of literals in the hypothesis).
An \emph{optimal} non-noisy solution minimises the cost function:

\begin{definition}[\textbf{Optimal non-noisy solution}]
\label{def:opthyp}
Given a LFF input $(E, B, \mathcal{H}, C, cost)$, a hypothesis $h \in \mathcal{H}_{C}$ is an \emph{optimal} non-noisy solution when (i) $h$ is a non-noisy solution, and (ii) $\forall h' \in \mathcal{H}_{C}$, where $h'$ is a non-noisy solution, $cost(h) \leq cost(h')$.
\end{definition}

\subsection{Noisy Learning From Failures}

A non-noisy solution must entail all the positive and none of the negative examples.
To tolerate noise, we relax this requirement.
We generalise a LFF input to allow a cost function to also take as input background knowledge $B$ and examples $E$, i.e. cost functions of the type $cost_{B,E} : \mathcal{H} \mapsto \mathbb{N}$.
In our noisy setting, any hypothesis $h \in \mathcal{H}$ is a noisy solution.
An \emph{optimal} noisy solution minimises the cost function:
\begin{definition}[\textbf{Optimal noisy solution}]
\label{def:opt_noisy_solution}
Given a noisy input $(E, B, \mathcal{H}, C, cost_{B,E})$, a hypothesis $h \in \mathcal{H}_{C}$ is an \emph{optimal} noisy solution when $\forall h' \in \mathcal{H}_{C}$, $cost_{B,E}(h) \leq cost_{B,E}(h')$.
\end{definition}

\subsection{Minimal Description Length}
Our noisy LFF setting generalises the LFF setting to allow for different cost functions.
A challenge in machine learning is choosing a suitable cost function.
According to complexity-based induction, the best hypothesis is the one that minimises the number of bits required to communicate the examples \cite{complexity}.
This concept corresponds to the hypothesis with minimal description complexity \cite{mdl}\footnote{Selecting an MDL hypothesis is equivalent to selecting a hypothesis with the maximum Bayes’ posterior probability \cite{mugg:ilp94}.}, where the idea is to trade off the complexity of a hypothesis (its size) with the fit to the data (training accuracy).

We use MDL as our cost function.
To define it, we use the terminology of \citet{complexity}.
The MDL principle states that the most probable hypothesis $h$ for the data $E$ is the one that minimises the complexity $L(h|E)$ of the hypothesis given the data.
The MDL principle can be expressed as finding a hypothesis that minimises $L(h)+L(E|h)$, where $L(h)$ is the syntactic complexity of a hypothesis $h$ and $L(E|h)$ is the complexity of the examples when coded using $h$.
We evaluate $L(h)$ with the function $size: \mathcal{H} \mapsto \mathbb{N}$, which measures the size of a hypothesis $h$ as the number of literals in it.
In a probabilistic setting, $L(E|h)$ is the log-likelihood of the data with respect to the hypothesis $h$.
However, there is debate about how to interpret $L(E|h)$ in a logical setting.
For instance, \citet{prooflength} use an encoding based on Turing machines (a proof complexity measure).
We evaluate $L(E|h)$ as the cost of sending the exceptions to the hypothesis, i.e. the number of false positives $fp_{E,B}(h)$ (simply $fp(h)$) and false negatives $fn_{E,B}(h)$ (simply $fn(h)$).
We define our MDL cost function:
\begin{definition}[\textbf{MDL cost function}]
\label{mdlcostfunction}
Given examples $E$ and background knowledge $B$, the MDL cost of a hypothesis $h\in \mathcal{H}$ is
$cost_{B,E}(h)=size(h) + fn_{E,B}(h) + fp_{E,B}(h)$.
\end{definition}

\noindent
In other words, the MDL cost of a hypothesis $h$ is the number of literals in $h$ plus the number of false positives and false negatives of $h$ on the training data.

In the rest of the paper, any reference to an \emph{optimal noisy solution} refers to an
optimal noisy solution with our MDL cost function.

\subsection{Noisy Constraints}
\label{constraints}
A LFF learner builds constraints from failures to restrict the hypothesis space.
The existing constraints for LFF are intolerant to noise.
For instance, if a hypothesis $h$ is inconsistent, a non-noisy generalisation constraint prunes generalisations of $h$ as they are also inconsistent.
However, in a noisy setting, a generalisation of $h$ might have a lower MDL cost.
Therefore, the existing constraints can prune optimal noisy solutions from the hypothesis space.

To overcome this limitation, we introduce constraints that tolerate noise.
These constraints are optimally sound for the noisy setting because they do not prune optimal noisy solutions from the hypothesis space.
Due to space limitations, we only describe one specialisation and one generalisation constraint.
The appendix contains a description of three other constraints.
All the proofs are in the appendix.


Let $h_1$ be a hypothesis with $tp(h_1)$ true positives and $h_2$ be a specialisation of $h_1$. Then $h_2$ has at most $tp(h_1)$ true positives.
Therefore, if $size(h_2)>tp(h_1)$ then the size of $h_2$ is greater than the number of positive examples it covers so $h_2$ cannot be in an optimal noisy solution:
\begin{proposition}[\textbf{Noisy specialisation constraint}] \label{specialisation}
Let $h_1$ be a hypothesis, $h_2$ be a specialisation of $h_1$, and $size(h_2)>tp(h_1)$.
Then $h_2$ cannot be in an optimal noisy solution.
\end{proposition}

\noindent
Similarly, let $h_1$ be a hypothesis with $fp(h_1)$ false positives and $h_2$ be a generalisation of $h_1$.
Then $h_2$ has at least $fp(h_1)$ false positives and a cost of at least $fp(h_1)+size(h_2)$.
We show that the cost of $h_2$ is greater than the cost of the empty hypothesis when $size(h_2) \geq |E^+|-fp(h_1)$:
\begin{proposition}[\textbf{Noisy generalisation constraint}] \label{generalisation}
Let $h_1$ be a hypothesis, $h_2$ be a generalisation of $h_1$, and $size(h_2) \geq |E^+|-fp(h_1)$.
Then $h_2$ cannot be in an optimal noisy solution.
\end{proposition}

\noindent
In the next section, we introduce \name{} which uses these optimally sound noisy constraints to learn programs.

\section{Algorithm}
\label{sec:impl}
We now describe our \name{} algorithm.
To explain our approach, we first describe \popper{} \cite{popper,combo}, which \name{} builds on.

\paragraph{\popper{}.}
\popper{} takes as input background knowledge, positive and negative training examples, and a maximum hypothesis size.
\popper{} starts with an ASP program $\mathcal{P}$.
Each model (answer set) of $\mathcal{P}$ corresponds to a hypothesis (a definite program).
\popper{} uses a generate, test, combine, and constrain loop to find a textually minimal non-noisy solution.
In the generate stage, \popper{} uses Clingo \cite{clingo}, an ASP system, to search for a model of $\mathcal{P}$ for increasing hypothesis sizes.
If there is no model, \popper{} increments the hypothesis size and loops again.
If there is a model, \popper{} converts it to a hypothesis $h$.
In the test stage, \popper{} uses Prolog to test $h$ on the examples.
If $h$ is a non-noisy solution, \popper{} returns it.
If $h$ covers at least one positive example and no negative examples, \popper{} adds $h$ to a set of promising programs.
In the combine stage, \popper{} searches for a combination (a union) of promising programs that covers all the positive examples and is minimal in size.
If \popper{} finds a combination, it sets the combination as the best solution so far and updates the maximum hypothesis size.
In the constrain stage, \popper{} uses $h$ to build hypothesis constraints (represented as ASP constraints).
\popper{} adds these constraints to $\mathcal{P}$ to prune models and thus prune the hypothesis space.
For instance, if $h$ is inconsistent, \popper{} builds a generalisation constraint to prune the generalisations of $h$ from the hypothesis space.
\popper{} repeats this loop until it finds a textually minimal non-noisy solution or there are no more hypotheses to test.

\begin{algorithm}[t]
\small
{
\begin{myalgorithm}[]
def maxsynth(bk, pos, neg):
  cons, promising, best_solution = {}, {}, {}
  size, max_mdl = 1, len(pos)
  while size $\leq$ max_mdl:
    h = generate(cons, size)
    if h == UNSAT:
      size += 1
      continue
    tp, fn, fp = test(pos, neg, bk, h)
    h_mdl = fn+fp+size(h)
    if h_mdl < max_mdl:
      best_solution = h
      max_mdl = h_mdl-1
    if tp>0 and not_rec(h) and not_pi(h):
      promising += h
      combination = combine(promising, max_mdl)
      if combination != UNSAT:
        best_solution = combination
        tp, fn, fp = test(pos, neg, bk, combination)
        max_mdl = fn+fp+size(combination)-1
    cons += constrain(h, fn, fp)
  return best_solution
\end{myalgorithm}
\caption{
\name{}
}
\label{alg:noisypopper}
}
\end{algorithm}

\subsection*{\name{}}
\name{} (Algorithm \ref{alg:noisypopper}) is similar to \popper{} except for a few key differences.
\popper{} returns the smallest hypothesis that entails all the positive and none of the negative examples, i.e. it is intolerant to noisy data.
By contrast, \name{} returns an MDL hypothesis, i.e. it is tolerant to noisy data.
To find an MDL hypothesis, \name{} differs by (i) also saving inconsistent programs as promising programs, (ii) finding an MDL combination in the combine stage, and (iii) using noise-tolerant constraints to prune non-MDL programs.
We describe these differences in turn.

\subsubsection{Promising Programs}
\popper{} only saves consistent programs as promising programs.
\popper{} is, therefore, intolerant to false negative training examples.
To handle noise, \name{} relaxes this requirement.
If a program $h$ covers at least one positive example, \name{} saves $h$ as a promising program (line 15), even if $h$ is inconsistent.
\name{} does not save a program if it is recursive or has predicate invention.
The reason is that a combination of recursive programs or programs with invented predicates can cover more examples than the union of the examples covered by each individual program.
For instance, consider the examples $\{f([1,3]), f([3,0]), f([3,1])\}$ and the hypotheses $h_1$ and $h_2$:
\[
    \begin{array}{l}
    h_1=\left\{
    \begin{array}{l}
\emph{f(A) $\leftarrow$ head(A,1)}\\
    \end{array}
    \right\}\\
    h_2=\left\{
    \begin{array}{l}
\emph{f(A) $\leftarrow$ head(A,0)}\\
\emph{f(A) $\leftarrow$ tail(A,B),f(B)}\\
    \end{array}
     \right\}
    \end{array}
\]
The hypothesis $h_1$ covers the first example and $h_2$ covers the second example but the hypothesis $h_1 \cup h_2$ covers all three examples.
Therefore, in the combine stage, we cannot simply reason about the coverage of a combination of programs using the union of coverage of the individual programs in the combination.
However, \name{} can learn MDL programs with recursion or predicate invention as they can be output by the generate stage and evaluated (lines 11-13).

\subsubsection{Combine}
In the combine stage, \popper{} searches for a combination of promising programs that covers all the positive examples and is minimal in size.
By contrast, \name{} searches for a combination of promising programs with MDL cost (line 16).
The initial maximum MDL cost is the number of positive examples which is the cost of the empty hypothesis.
If we find a combination in the combine stage, we update the maximum MDL cost (line 20).

We formulate the search for an MDL combination of programs as a MaxSAT problem~\cite{maxsat}.
In MaxSAT, given a set of hard clauses and a set of soft clauses with an associated weight, the task is to find a truth assignment which satisfies each hard clause and minimises the sum of the weights of falsified soft clauses.

Our MaxSAT encoding is as follows.
For each promising program $h$, we use a variable $p_h$ to indicate whether $h$ is in the combination.
For each example $e \in E^+ \cup E^-$, we use a variable $c_e$ to indicate whether the combination covers $e$.
For each positive example $e \in E^+$, we include the hard clause $c_e \rightarrow \bigvee_{B \cup h \models e} p_h$ to ensure that, if the combination covers $e$, then at least one of the programs in the combination covers $e$.
For each negative example $e \in E^-$, we include the hard clause $\neg c_e \rightarrow \bigwedge_{B \cup h \models e} \neg p_h$ to ensure that, if the combination does not cover $e$, then none of the programs in the combination covers $e$.
We encode the MDL cost function as follows.
For each promising program $h$ we include the soft clause $(\neg p_h)$ with weight $size(h)$.
For each positive example $e \in E^+$, we include the soft clause $(c_e)$ with weight $1$.
For each negative example $e \in E^-$, we include the soft clause $(\neg c_e)$ with weight $1$.
We use a MaxSAT solver on this encoding.
The MaxSAT solver finds an optimal solution which corresponds to a combination of promising programs that minimises the MDL cost function.

\subsubsection{Constrain}
In the constrain stage (line 21), \name{} uses our optimally sound constraints (Section \ref{constraints}) to prune the hypothesis space. For instance, given a hypothesis $h_1$, \name{} prunes all generalisations of $h_1$ with size greater than $|E^+|-fp(h_1)$ (Proposition \ref{generalisation}). By contrast, \popper{} prunes all generalisations of an inconsistent hypothesis.

\subsubsection{Correctness}
We show that \name{} returns an optimal noisy solution.
\begin{theorem}[\textbf{Correctness}]\label{correctness}
\name{} returns an optimal noisy solution if one exists.
\end{theorem}

\noindent
\textbf{Proof.}
\emph{The proof is in the appendix.
We first show that \name{} without any noisy constraints returns an optimal noisy hypothesis, and then
that our noise-tolerant constraints are optimally sound (Propositions \ref{specialisation} and \ref{generalisation}).
}

\section{Experiments}
To test our claim that \name{} can learn programs from noisy data, our experiments aim to answer the question:

\begin{enumerate}
\item[\textbf{Q1}] Can \name{} learn programs from noisy data?
\end{enumerate}
To answer \textbf{Q1}, we evaluate \name{} on a variety of tasks with noisy data.
We compare \name{} against \ale{} \cite{aleph}, \popper{}, and \noisypopper{} \cite{noisypopper}\footnote{We considered other systems. Rule selection approaches \cite{aspal,dilp,hexmil} precompute every possible rule which is infeasible on our datasets.
Metarule-based approaches \cite{mugg:metagold} are unusable in practice \cite{ilp30}.
Rule learning systems \cite{amie+} can only use unary and binary relations but our experiments need relations with arity greater than two.
}.
We use these systems because they can learn definite recursive programs.
\ale{} is a set covering approach that supports noise.
\noisypopper{} can handle noisy data but can only learn small programs.
Because of space limitations and its poor performance, the results for \noisypopper{} are in the appendix.

To evaluate how \name{} handles different amounts of noise, our experiments aim to answer the question:
\begin{enumerate}
\item[\textbf{Q2}] How well does \name{} handle progressively more noise?
\end{enumerate}
To answer \textbf{Q2}, we evaluate the performance of \name{} on domains where we can progressively increase the amount of noise.
For an increasing noise amount $p$, we randomly change the label of a proportion $p$ of the training examples.

We claim that our noisy constraints (Section \ref{constraints}) can improve learning performance by pruning non-MDL programs from the hypothesis space.
To evaluate this claim, our experiments aim to answer the question:
\begin{enumerate}
\item[\textbf{Q3}] Can noisy constraints reduce learning times compared to unconstrained learning?
\end{enumerate}
To answer \textbf{Q3}, we compare the learning time of \name{} with and without noisy constraints.

Our approach should improve learning performance when learning programs from noisy data.
However, it is often unknown whether the data is noisy.
To evaluate the overhead of handling noise, our experiments aim to answer the question:
\begin{enumerate}
\item[\textbf{Q4}] What is the overhead of \name{} on noiseless problems?
\end{enumerate}
To answer \textbf{Q4}, we compare the performance of \name{} and \popper{} on standard benchmarks which are not noisy.

\subsubsection{Domains}

We briefly describe our five domains.
The appendix includes more details.

\textbf{IGGP.}
The goal of \emph{inductive general game playing} \cite{iggp} (IGGP) is to induce rules to explain game traces from the general game playing competition \cite{ggp}.

\textbf{Program synthesis.}
We use a program synthesis dataset \cite{popper}.
These tasks are list transformation tasks which involve learning recursive programs.

\textbf{Zendo.}
Zendo is an inductive game where the goal is to find a rule by building structures of pieces.
The game interests cognitive scientists \cite{zendo}.

\textbf{Alzheimer.}
These real-world tasks \cite{alzheimer} involve learning rules describing four properties desirable for drug design against Alzheimer's disease.

\textbf{Wn18RR.} Wn18rr \cite{wn18rr} is a real-world knowledge base with 11 relations from WordNet.

\subsubsection{Systems}
\name{}, \popper{}, and \noisypopper{} use identical biases so the comparison between them is fair.
\name{} uses the UWrMaxSat solver \cite{uwrmaxsat} in the combine stage.
To perform a direct comparison, we modify \popper{} to also use the UWrMaxSat solver in its combine stage.
We use the default cost function (coverage) for \ale{}.
We have tried to make a fair comparison with \ale{} but, since it has many additional settings, it is naturally plausible that further parameter tuning could improve its performance \cite{ashwinsettings}.
The appendix contains more details about the systems.

\subsubsection{Experimental Setup}
We measure predictive accuracy and learning time given a maximum learning time of 20 minutes.
We repeat all the experiments 10 times and calculate the mean and standard error. We use an 8-Core 3.2 GHz Apple M1 and a single CPU.


\subsection*{Experimental Results}

\subsubsection{Experiment 1: Comparison against SOTA}
Table \ref{tab:accuracies} shows the predictive accuracies of the systems on the datasets.
It shows that \name{} (i) consistently achieves high accuracy on most tasks, and (ii) comprehensively outperforms existing systems in terms of predictive accuracy.
A paired t-test shows \name{} significantly ($p < 0.01$) outperforms \popper{} on 25/42 tasks, achieves similar accuracies on 13/42 tasks, and is significantly outperformed by \popper{} on 4/42 tasks.
For instance, \name{} has high accuracy (at least 94\%) on all \emph{zendo} tasks while \popper{} struggles when there is noise.
While \popper{} searches for a hypothesis that entails all the positive and no negative examples, \name{} tolerates misclassified examples.

\name{} outperforms \ale{} on the recursive tasks because \ale{} struggles to learn recursive programs.
\name{} also outperforms \ale{} on some non-recursive tasks.
For instance, on \emph{iggp-coins}, \name{} achieves 100\% predictive accuracy on the testing examples even with 20\% noise in the training examples.
One reason is that \ale{} does not consider the size of a hypothesis in its (default) cost function and thus often overfits.
\ale{} also sometimes timeouts, such as on the \emph{wn18rr} tasks, in which case it does not return any hypothesis.

\begin{table}[ht!]
\centering
\footnotesize
\begin{tabular}{@{}l|ccc@{}}
\textbf{Task} & \textbf{\name} & \textbf{\popper} & \textbf{\ale}\\
\midrule
\emph{iggp-md (0)} & 75 $\pm$ 0 & \textbf{100 $\pm$ 0} & \textbf{100 $\pm$ 0}\\
\emph{iggp-md (10)} & 65 $\pm$ 4 & 76 $\pm$ 6 & \textbf{85 $\pm$ 6}\\
\emph{iggp-md (20)} & 58 $\pm$ 4 & 68 $\pm$ 6 & \textbf{73 $\pm$ 7}\\
\emph{iggp-buttons (0)} & 80 $\pm$ 0 & \textbf{100 $\pm$ 0} & \textbf{100 $\pm$ 0}\\
\emph{iggp-buttons (10)} & \textbf{79 $\pm$ 1} & \textbf{79 $\pm$ 3} & 50 $\pm$ 0\\
\emph{iggp-buttons (20)} & \textbf{77 $\pm$ 1} & 63 $\pm$ 1 & 50 $\pm$ 0\\
\emph{iggp-coins (0)} & \textbf{100 $\pm$ 0} & \textbf{100 $\pm$ 0} & 50 $\pm$ 0\\
\emph{iggp-coins (10)} & \textbf{100 $\pm$ 0} & 54 $\pm$ 1 & 50 $\pm$ 0\\
\emph{iggp-coins (20)} & \textbf{100 $\pm$ 0} & 50 $\pm$ 0 & 50 $\pm$ 0\\
\emph{iggp-rps (0)} & \textbf{100 $\pm$ 0} & \textbf{100 $\pm$ 0} & \textbf{100 $\pm$ 0}\\
\emph{iggp-rps (10)} & \textbf{100 $\pm$ 0} & 63 $\pm$ 2 & 73 $\pm$ 3\\
\emph{iggp-rps (20)} & \textbf{100 $\pm$ 0} & 59 $\pm$ 2 & 50 $\pm$ 0\\
\midrule
\emph{zendo1 (0)} & 99 $\pm$ 0 & \textbf{100 $\pm$ 0} & \textbf{100 $\pm$ 0}\\
\emph{zendo1 (10)} & \textbf{99 $\pm$ 0} & 82 $\pm$ 1 & 82 $\pm$ 1\\
\emph{zendo1 (20)} & \textbf{99 $\pm$ 0} & 74 $\pm$ 1 & 76 $\pm$ 1\\
\emph{zendo2 (0)} & \textbf{100 $\pm$ 0} & \textbf{100 $\pm$ 0} & \textbf{100 $\pm$ 0}\\
\emph{zendo2 (10)} & \textbf{100 $\pm$ 0} & 64 $\pm$ 1 & 65 $\pm$ 1\\
\emph{zendo2 (20)} & \textbf{100 $\pm$ 0} & 58 $\pm$ 1 & 60 $\pm$ 1\\
\emph{zendo3 (0)} & 98 $\pm$ 0 & \textbf{99 $\pm$ 0} & \textbf{99 $\pm$ 0}\\
\emph{zendo3 (10)} & \textbf{98 $\pm$ 0} & 72 $\pm$ 1 & 72 $\pm$ 1\\
\emph{zendo3 (20)} & \textbf{97 $\pm$ 1} & 68 $\pm$ 1 & 70 $\pm$ 1\\
\emph{zendo4 (0)} & 98 $\pm$ 0 & \textbf{99 $\pm$ 0} & \textbf{99 $\pm$ 0}\\
\emph{zendo4 (10)} & \textbf{96 $\pm$ 0} & 81 $\pm$ 1 & 82 $\pm$ 1\\
\emph{zendo4 (20)} & \textbf{94 $\pm$ 1} & 74 $\pm$ 1 & 77 $\pm$ 1\\
\midrule
\emph{dropk (0)} & \textbf{100 $\pm$ 0} & \textbf{100 $\pm$ 0} & 55 $\pm$ 5\\
\emph{dropk (10)} & \textbf{100 $\pm$ 0} & 54 $\pm$ 1 & 50 $\pm$ 0\\
\emph{dropk (20)} & \textbf{100 $\pm$ 0} & 54 $\pm$ 1 & 50 $\pm$ 0\\
\emph{evens (0)} & \textbf{100 $\pm$ 0} & \textbf{100 $\pm$ 0} & 57 $\pm$ 3\\
\emph{evens (10)} & \textbf{100 $\pm$ 0} & 52 $\pm$ 1 & 52 $\pm$ 1\\
\emph{evens (20)} & \textbf{100 $\pm$ 0} & 51 $\pm$ 0 & 51 $\pm$ 0\\
\emph{reverse (0)} & \textbf{100 $\pm$ 0} & \textbf{100 $\pm$ 0} & 50 $\pm$ 0\\
\emph{reverse (10)} & \textbf{100 $\pm$ 0} & 52 $\pm$ 0 & 50 $\pm$ 0\\
\emph{reverse (20)} & \textbf{100 $\pm$ 0} & 52 $\pm$ 0 & 50 $\pm$ 0\\
\emph{sorted (0)} & \textbf{100 $\pm$ 0} & \textbf{100 $\pm$ 0} & 68 $\pm$ 5\\
\emph{sorted (10)} & \textbf{100 $\pm$ 0} & 64 $\pm$ 2 & 63 $\pm$ 2\\
\emph{sorted (20)} & \textbf{100 $\pm$ 0} & 58 $\pm$ 1 & 56 $\pm$ 2\\
\midrule
\emph{alzheimer\_acetyl} & \textbf{68 $\pm$ 1} & 56 $\pm$ 0 & 50 $\pm$ 0\\
\emph{alzheimer\_amine} & \textbf{76 $\pm$ 1} & 69 $\pm$ 1 & 73 $\pm$ 1\\
\emph{alzheimer\_mem} & \textbf{63 $\pm$ 1} & 51 $\pm$ 0 & 61 $\pm$ 1\\
\emph{alzheimer\_toxic} & 74 $\pm$ 1 & 64 $\pm$ 1 & \textbf{83 $\pm$ 1}\\
\midrule
\emph{wn18rr1} & \textbf{98 $\pm$ 0} & 95 $\pm$ 1 & 50 $\pm$ 0\\
\emph{wn18rr2} & \textbf{79 $\pm$ 1} & 78 $\pm$ 1 & 50 $\pm$ 0\\
\end{tabular}
\caption{
Predictive accuracies.
Numbers in parentheses indicate the amount of noise added.
The amount of noise is unknown when unspecified.
}
\label{tab:accuracies}
\end{table}

\name{} does not always achieve 100\% predictive accuracy despite learning an MDL hypothesis, such as on the \emph{iggp-md} tasks.
The reason is that an MDL hypothesis is not necessarily the hypothesis with the highest predictive accuracy \cite{pedro:occam,filip:occam}.

\name{} and \popper{} are anytime systems.
If the search time exceeds a timeout, \name{} and \popper{} return the best hypothesis found thus far.
\name{} terminates on all \emph{iggp}, \emph{zendo}, and \emph{alzheimer} tasks, which means it learns an MDL solution.
\name{} returns the best solution found within timeout for most \emph{program synthesis} tasks and \emph{wn18rr2}.

To understand how accuracy varies with learning time, we set a timeout of $t$ seconds for increasing values of $t$.
Figures \ref{fig:timealzheimer} and \ref{fig:timezendo} show the predictive accuracies of the best hypothesis found when increasing the timeout.
This result shows that \name{} can often quickly find an optimal (MDL) hypothesis.
For instance, on \emph{alzheimer-toxic}, \name{} takes only 13s to find an optimal hypothesis but needs 48s to prove that this hypothesis is optimal.
Likewise, on \emph{zendo2 (20)}, \name{} takes only 60s to find an optimal hypothesis but needs 151s more to prove this hypothesis is optimal.

Overall, these results suggest that the answer to \textbf{Q1} is that \name{} can (i) learn programs, including recursive programs, with high accuracy from noisy data, and (ii) outperform existing systems in terms of predictive accuracies.

\begin{figure}[t]
  \begin{minipage}{0.05\textwidth}
\begin{tikzpicture}
\begin{customlegend}[legend columns=5,legend style={nodes={scale=1, transform shape},align=left,column sep=0ex},
        legend entries={\name, \ale, \popper}]
        \addlegendimage{red,mark=diamond*}
        \addlegendimage{blue,mark=square*}
        \addlegendimage{green,mark=triangle*}
\end{customlegend}
\end{tikzpicture}
\end{minipage}\hfill\\
 \begin{minipage}{0.23\textwidth}
\resizebox{\columnwidth}{!}{
\pgfplotstableread[col sep=comma]{./results/alzheimer_toxic_maxsynth_maxsat_c_timeout.csv}\alzheimermaxsynth
\pgfplotstableread[col sep=comma]{./results/alzheimer_toxic_popper_timeout.csv}\alzheimerpopper
\pgfplotstableread[col sep=comma]{./results/alzheimer_toxic_aleph_False_timeout.csv}\alzheimeraleph
\begin{tikzpicture}
\begin{axis}[
  legend style={at={(0.5,0.35)},anchor=west},
   legend style={font=\normalsize},
  tick label style={font=\huge},
  xlabel style={font=\huge},
  ylabel style={font=\huge},
  ytick={50, 60, 70, 80, 90, 100},
  xtick={0, 10, 20, 30, 40, 50},
  xmin=0,xmax=45,
  xlabel=Timeout (s),
  ylabel=Accuracy (\%),
label style={font=\huge},
  ]
\addplot+[red,mark=diamond*,
                error bars/.cd,
                y dir=both,
                error mark,
                y explicit]table[x=timeout,y=acc_av,y error=acc_std] {\alzheimermaxsynth};
\addplot+[green,mark=triangle*,
                error bars/.cd,
                y dir=both,
                error mark,
                y explicit]table[x=timeout,y=acc_av,y error=acc_std] {\alzheimerpopper};
\addplot+[blue,mark=square*,
                error bars/.cd,
                y dir=both,
                error mark,
                y explicit]table[x=timeout,y=acc_av,y error=acc_std] {\alzheimeraleph};

\end{axis}
\end{tikzpicture}}
\caption{Predictive accuracy with larger timeouts on \emph{alzheimer-toxic}.}
\label{fig:timealzheimer}
\end{minipage}\hfill
\begin{minipage}{0.23\textwidth}
\resizebox{\columnwidth}{!}{

\pgfplotstableread[col sep=comma]{./results/zendo2__0.2_maxsynth_maxsat_c_timeout.csv}\zendomaxsynth
\pgfplotstableread[col sep=comma]{./results/zendo2__0.2_popper_timeout.csv}\zendopopper
\pgfplotstableread[col sep=comma]{./results/zendo2__0.2_aleph_False_timeout.csv}\zendoaleph

\begin{tikzpicture}
\begin{axis}[
  legend style={at={(0.5,0.35)},anchor=west},
   legend style={font=\normalsize},
  tick label style={font=\huge},
 xlabel style={font=\huge},
 ylabel style={font=\huge},
  ytick={50, 60, 70, 80, 90, 100},
 xtick={0,60,120,180,240},
 xmin=0,xmax=240,
 xlabel=Timeout (s),
  ylabel=Accuracy (\%),
label style={font=\huge},
  ]
\addplot+[red,mark=diamond*,
                error bars/.cd,
                y dir=both,
                error mark,
                y explicit]table[x=timeout,y=acc_av,y error=acc_std] {\zendomaxsynth};
\addplot+[green,mark=triangle*,
                error bars/.cd,
                y dir=both,
                error mark,
                y explicit]table[x=timeout,y=acc_av,y error=acc_std] {\zendopopper};
\addplot+[blue,mark=square*,
                error bars/.cd,
                y dir=both,
                error mark,
                y explicit]table[x=timeout,y=acc_av,y error=acc_std] {\zendoaleph};

\end{axis}
\end{tikzpicture}}
\caption{Predictive accuracy with larger timeouts on \emph{zendo2 (20)}.}
\label{fig:timezendo}
\end{minipage}
\end{figure}

\subsubsection{Experiment 2: Noise Tolerance}
Figures \ref{fig:accbuttons} and \ref{fig:accdropk} show the predictive accuracies of the systems on two tasks when increasing the amount of noise.
These results show that the performance of \name{} degrades slower with increasing amounts of noise compared to \popper{}.
\name{} can scale to problems with up to 30\% of noise  while \popper{} struggles from 10\% of noise.
For instance, on \emph{iggp-rps} with 30\% of noise, \popper{} and \ale{} have less than 55\% accuracy, whereas \name{} has over 90\% accuracy.
\popper{} is not robust to false positives.
It returns a hypothesis which is consistent but only covers a fraction of the positive examples.
\ale{} typically overfits the data.
Overall, these results suggest that the answer to \textbf{Q2} is that \name{} can scale to moderate amounts of noise.

\pgfplotstableread[col sep=comma]{./results/iggp-scissors_paper_stone_next_maxsynth_maxsat_c_noise.csv}\rpsmaxsynth
\pgfplotstableread[col sep=comma]{./results/iggp-scissors_paper_stone_next_popper_noise.csv}\rpspopper
\pgfplotstableread[col sep=comma]{./results/iggp-scissors_paper_stone_next_aleph_False_noise.csv}\rpsaleph

\pgfplotstableread[col sep=comma]{./results/dropk_maxsynth_maxsat_c_noise.csv}\dropkmaxsynth
\pgfplotstableread[col sep=comma]{./results/dropk_popper_noise.csv}\dropkpopper
\pgfplotstableread[col sep=comma]{./results/dropk_aleph_False_noise.csv}\dropkaleph

\begin{figure}[ht!]
  \begin{minipage}{0.05\textwidth}
\begin{tikzpicture}
\begin{customlegend}[legend columns=5,legend style={nodes={scale=1, transform shape},align=left,column sep=0ex},
        legend entries={\name, \ale, \popper}]
        \addlegendimage{red,mark=diamond*}
        \addlegendimage{blue,mark=square*}
        \addlegendimage{green,mark=triangle*}
\end{customlegend}
\end{tikzpicture}
\end{minipage}\hfill\\
 \begin{minipage}{0.23\textwidth}
\resizebox{\columnwidth}{!}{
\begin{tikzpicture}
\begin{axis}[
  legend style={at={(0.5,0.35)},anchor=west},
   legend style={font=\normalsize},
  tick label style={font=\huge},
  xtick={0, 10, 20, 30, 40, 50},
  ytick={50, 60, 70, 80, 90, 100},
  xmin=0,xmax=40.1,
  xlabel=Noise amount (\%),
  ylabel=Accuracy (\%),
label style={font=\huge},
  ]
\addplot+[red,mark=diamond*,
                error bars/.cd,
                y dir=both,
                error mark,
                y explicit]table[x=xs,y=acc_av,y error=acc_std] {\rpsmaxsynth};
\addplot+[green,mark=triangle*,
                error bars/.cd,
                y dir=both,
                error mark,
                y explicit]table[x=xs,y=acc_av,y error=acc_std] {\rpspopper};
\addplot+[blue,mark=square*,
                error bars/.cd,
                y dir=both,
                error mark,
                y explicit]table[x=xs,y=acc_av,y error=acc_std] {\rpsaleph};
\end{axis}
\end{tikzpicture}}
\caption{Accuracy versus the noise amount on \emph{iggp-rps}.}
\label{fig:accbuttons}
\end{minipage}\hfill
 \begin{minipage}{0.23\textwidth}
\resizebox{\columnwidth}{!}{
\begin{tikzpicture}
\begin{axis}[
  legend style={at={(0.5,0.35)},anchor=west},
   legend style={font=\normalsize},
  tick label style={font=\huge},
  xtick={0, 10, 20, 30, 40, 50},
  ytick={50, 60, 70, 80, 90, 100},
  xmin=0,xmax=40.1,
  xlabel=Noise amount (\%),
  ylabel=Accuracy (\%),
label style={font=\huge},
  ]
  \addplot+[red,mark=diamond*,
                error bars/.cd,
                y dir=both,
                error mark,
                y explicit]table[x=xs,y=acc_av,y error=acc_std] {\dropkmaxsynth};
\addplot+[green,mark=triangle*,
                error bars/.cd,
                y dir=both,
                error mark,
                y explicit]table[x=xs,y=acc_av,y error=acc_std] {\dropkpopper};
\addplot+[blue,mark=square*,
                error bars/.cd,
                y dir=both,
                error mark,
                y explicit]table[x=xs,y=acc_av,y error=acc_std] {\dropkaleph};
\end{axis}
\end{tikzpicture}}
\caption{Accuracy versus the noise amount on \emph{dropk}.}
\label{fig:accdropk}
\end{minipage}
\end{figure}

\subsubsection{Experiment 3: Noisy Constraints}
\begin{table}[ht!]
\centering
\footnotesize
\begin{tabular}{@{}l|cccc@{}}
\textbf{Task} & \textbf{Without} & \textbf{With} & \textbf{Difference}\\
\midrule
\emph{iggp-md (0)} & 14 $\pm$ 0 & 1 $\pm$ 0& \textbf{-92\%}\\
\emph{iggp-md (10)} & 109 $\pm$ 2 & 2 $\pm$ 0& \textbf{-98\%}\\
\emph{iggp-md (20)} & 103 $\pm$ 1 & 2 $\pm$ 0& \textbf{-98\%}\\
\emph{iggp-buttons (0)} & 61 $\pm$ 0 & 7 $\pm$ 0& \textbf{-88\%}\\
\emph{iggp-buttons (10)} & 61 $\pm$ 1 & 9 $\pm$ 0& \textbf{-85\%}\\
\emph{iggp-buttons (20)} & 57 $\pm$ 0 & 10 $\pm$ 0& \textbf{-82\%}\\
\emph{iggp-coins (0)} & 615 $\pm$ 5 & 138 $\pm$ 1& \textbf{-77\%}\\
\emph{iggp-coins (10)} & 631 $\pm$ 14 & 141 $\pm$ 2& \textbf{-77\%}\\
\emph{iggp-coins (20)} & 596 $\pm$ 2 & 144 $\pm$ 2& \textbf{-75\%}\\
\emph{iggp-rps (0)} & 195 $\pm$ 1 & 50 $\pm$ 1& \textbf{-74\%}\\
\emph{iggp-rps (10)} & 197 $\pm$ 1 & 60 $\pm$ 2& \textbf{-69\%}\\
\emph{iggp-rps (20)} & 193 $\pm$ 1 & 66 $\pm$ 1& \textbf{-65\%}\\
\midrule
\emph{zendo1 (0)} & 33 $\pm$ 9 & 13 $\pm$ 3& \textbf{-60\%}\\
\emph{zendo1 (10)} & 648 $\pm$ 3 & 77 $\pm$ 4& \textbf{-88\%}\\
\emph{zendo1 (20)} & 688 $\pm$ 7 & 100 $\pm$ 13& \textbf{-85\%}\\
\emph{zendo2 (0)} & 603 $\pm$ 4 & 48 $\pm$ 1& \textbf{-92\%}\\
\emph{zendo2 (10)} & 611 $\pm$ 1 & 48 $\pm$ 3& \textbf{-92\%}\\
\emph{zendo2 (20)} & 766 $\pm$ 70 & 118 $\pm$ 36 & \textbf{-84\%}\\
\emph{zendo3 (0)} & 613 $\pm$ 2 & 49 $\pm$ 3& \textbf{-92\%}\\
\emph{zendo3 (10)} & 626 $\pm$ 2 & 62 $\pm$ 2& \textbf{-90\%}\\
\emph{zendo3 (20)} & 834 $\pm$ 66 & 190 $\pm$ 112& \textbf{-77\%}\\
\emph{zendo4 (0)} & 594 $\pm$ 4 & 43 $\pm$ 3& \textbf{-92\%}\\
\emph{zendo4 (10)} & 616 $\pm$ 3 & 58 $\pm$ 2& \textbf{-90\%}\\
\emph{zendo4 (20)} & 767 $\pm$ 36 & 122 $\pm$ 32& \textbf{-84\%}\\
\midrule
\emph{dropk (0)} & 541 $\pm$ 5 & 7 $\pm$ 1& \textbf{-98\%}\\
\emph{evens (0)} & 770 $\pm$ 2 & 7 $\pm$ 0& \textbf{-99\%}\\
\emph{reverse (0)} & \emph{timeout} & 45 $\pm$ 7& \textbf{-96\%}\\
\emph{sorted (0)} & 1182 $\pm$ 8 & 31 $\pm$ 3& \textbf{-97\%}\\
\midrule
\emph{alzheimer\_acetyl} & \emph{timeout} & 133 $\pm$ 5& \textbf{-88\%}\\
\emph{alzheimer\_amine} & \emph{timeout} & 73 $\pm$ 3& \textbf{-93\%}\\
\emph{alzheimer\_mem} & \emph{timeout} & 79 $\pm$ 3& \textbf{-93\%}\\
\emph{alzheimer\_toxic} & \emph{timeout} & 61 $\pm$ 6& \textbf{-94\%}\\
\midrule
\emph{wn18rr1} & \emph{timeout} & 534 $\pm$ 14 & \textbf{-55\%}\\
\end{tabular}
\caption{Learning time for \name{} with and without noisy constraints. We show tasks where approaches differ. The full table is in the appendix.
}
\label{tab:constraints_time}
\end{table}

Table \ref{tab:constraints_time} shows the learning times of \name{} with and without noisy constraints (Section \ref{constraints}).
It shows that our constraints can drastically reduce learning times.
A paired t-test confirms the significance of the difference for all tasks ($p<0.01$).
The appendix shows the predictive accuracies, which are equal or higher with noisy constraints.
This result shows that our noisy constraints are highly effective at soundly pruning the hypothesis space.
For instance, on \emph{iggp-md (10)}, \name{} considers 21,025 programs without constraints.
By contrast, with constraints, \name{} considers only 136 programs, a 99\% reduction. 
Similarly, \name{} considers 176,453 programs for \emph{zendo2 (20)} without constraints and only 5,503 programs with constraints, a 97\% reduction.
The overhead of analysing hypotheses and imposing constraints is small.
For instance, on \emph{iggp-md (10)}, \name{} spends 0.7s building constraints but this pruning reduces the total learning time from 109s to 2s. Overall, these results suggest that the answer to \textbf{Q3} is that noisy constraints can drastically reduce learning times.

\subsubsection{Experiment 4: Overhead}
Table \ref{tab:accuracies} shows the predictive accuracies of \name{} and \popper{} on noiseless problems.
These results show that \name{} often can find a non-noisy solution.
However, \name{} may return a simpler hypothesis than \popper{}.
For instance, on \emph{iggp-md (0)}, \name{} returns a hypothesis of size 5.
This hypothesis misclassifies 2 training positive examples and therefore has a cost of 7. Its predictive accuracy is 75\%.
By contrast, \popper{} finds a hypothesis of size 11 with maximal predictive accuracy (100\%) on the test data.
As \citet{vitanyi2000minimum} discuss, MDL interprets perfect data as data obtained from a simpler hypothesis subject to measuring errors.

Table \ref{tab:time_nonoise} shows the learning times.
It shows that \name{} often has similar learning times to \popper{}.
For instance, both systems require 7s on \emph{iggp-buttons~(0)} and around 50s on \emph{iggp-rps~(0)}.
\name{} can be faster than \popper{}.
A paired t-test shows \name{} significantly outperforms \popper{} on 7/12 tasks ($p<0.01$).
For instance, on \emph{zendo2~(0)}, \name{} takes 48s whilst \popper{} takes 102s.
The pruning by \name{} can be effective.
For instance, given any hypothesis $h$, \name{} prunes specialisations of size greater than $fp(h)$, whereas \popper{} only prunes specialisations of consistent hypotheses.
Also, \name{} sometimes returns a smaller hypothesis than \popper{} and thus searches up to a smaller depth, such as for \emph{iggp-md~(0)}.

Overall, these results suggest that the answer to \textbf{Q4} is that unnecessarily tolerating noise is not prohibitively expensive and often leads to similar performance.

\begin{table}[t]
\centering
\footnotesize
\begin{tabular}{@{}l|cc@{}}
\textbf{Task} & \textbf{\name} & \textbf{\popper}\\
\midrule
\emph{iggp-md (0)} & \textbf{1 $\pm$ 0} & 11 $\pm$ 0\\
\emph{iggp-buttons (0)} & \textbf{7 $\pm$ 0} & \textbf{7 $\pm$ 0}\\
\emph{iggp-coins (0)} & \textbf{138 $\pm$ 1} & 147 $\pm$ 2\\
\emph{iggp-rps (0)} & \textbf{50 $\pm$ 1} & 53 $\pm$ 1\\
\midrule
\emph{zendo1 (0)} & \textbf{13 $\pm$ 3} & 23 $\pm$ 4\\
\emph{zendo2 (0)} & \textbf{48 $\pm$ 1} & 102 $\pm$ 2\\
\emph{zendo3 (0)} & \textbf{49 $\pm$ 3} & 85 $\pm$ 3\\
\emph{zendo4 (0)} & \textbf{43 $\pm$ 3} & 79 $\pm$ 4\\
\midrule
\emph{dropk (0)} & 7 $\pm$ 1 & \textbf{4 $\pm$ 1}\\
\emph{evens (0)} & \textbf{7 $\pm$ 0} & 8 $\pm$ 0\\
\emph{reverse (0)} & \textbf{45 $\pm$ 7} & 65 $\pm$ 9\\
\emph{sorted (0)} & 31 $\pm$ 3 & \textbf{19 $\pm$ 2}\\
\end{tabular}
\caption{
Learning times on non-noisy tasks.
}
\label{tab:time_nonoise}
\end{table}

\section{Conclusions and Limitations}
We have introduced an ILP approach that learns MDL programs from noisy examples, including recursive programs.
Our approach first learns small programs that generalise a subset of the positive examples and then combines them to build an MDL program.
We implemented our idea in \name{}, which uses a MaxSAT solver to find an MDL combination of programs.
Our empirical results on multiple domains show that \name{} can (i) substantially improve predictive accuracies compared to other systems, and (ii) scale to moderate amounts of noise (30\%).
Our results also show that our noisy constraints can reduce learning times by 99\%.
Overall, this paper shows that \name{} can learn accurate hypotheses for noisy problems that other systems cannot.

\paragraph{Limitations.}
We use MDL as our criterion for optimality.
Our experiments show that an MDL hypothesis does not necessarily have the lowest generalisation error, as discussed by \citet{pedro:occam}.
To overcome this limitation, future work should investigate alternative cost functions \cite{nada:metrics}.
For instance, \citet{hernandez2000explanatory} discuss creative alternatives to MDL.

\bibliography{manuscript}

\begin{appendices}
 
\section{Terminology}
\label{sec:bk}
\subsection{Logic Programming}
We assume familiarity with logic programming \cite{lloyd:book} but restate some key relevant notation. A \emph{variable} is a string of characters starting with an uppercase letter. A \emph{predicate} symbol is a string of characters starting with a lowercase letter. The \emph{arity} $n$ of a function or predicate symbol is the number of arguments it takes. An \emph{atom} is a tuple $p(t_1, ..., t_n)$, where $p$ is a predicate of arity $n$ and $t_1$, ..., $t_n$ are terms, either variables or constants. An atom is \emph{ground} if it contains no variables. A \emph{literal} is an atom or the negation of an atom. A \emph{clause} is a set of literals. A \emph{clausal theory} is a set of clauses. A \emph{constraint} is a clause without a positive literal. A \emph{definite} clause is a clause with exactly one positive literal. 
A \emph{hypothesis} is a set of definite clauses.
We use the term \emph{program} interchangeably with hypothesis, i.e. a \emph{hypothesis} is a \emph{program}.
A \emph{substitution} $\theta = \{v_1 / t_1, ..., v_n/t_n \}$ is the simultaneous replacement of each variable $v_i$ by its corresponding term $t_i$. 
A clause $c_1$ \emph{subsumes} a clause $c_2$ if and only if there exists a substitution $\theta$ such that $c_1 \theta \subseteq c_2$. 
A program $h_1$ subsumes a program $h_2$, denoted $h_1 \preceq h_2$, if and only if $\forall c_2 \in h_2, \exists c_1 \in h_1$ such that $c_1$ subsumes $c_2$. A program $h_1$ is a \emph{specialisation} of a program $h_2$ if and only if $h_2 \preceq h_1$. A program $h_1$ is a \emph{generalisation} of a program $h_2$ if and only if $h_1 \preceq h_2$.

\subsection{ILP Terminology}
Let $(E, B, \mathcal{H}, C, cost)$ be a LFF input where $E=(E^{+}, E^{-})$ and $h \in \mathcal{H}$ be a hypothesis. A false negative for the hypothesis $h$ is a positive example uncovered by the hypothesis $h$. The number of false negative of $h$, represented as $fn_{E,B}(h)$, or simply $fn(h)$, is:
\begin{align*}
fn_{E,B}(h) = |\{e \in E^+ \mid \; B \cup h \not\models e\}|
\end{align*}
A false positive for the hypothesis $h$ is a negative example covered by the hypothesis $h$. The number of false positive of $h$, represented as $fp_{E,B}(h)$, or simply $fp(h)$, is:
\begin{align*}
fp_{E,B}(h) = |\{e \in E^- \mid \; B \cup h \models e\}|
\end{align*}

\noindent
A hypothesis $h$ is \emph{incomplete} when $\exists e \in E^+, \; h \cup B \not \models e$ i.e. when $fn(h)>0$. 
A hypothesis $h$ is \emph{inconsistent} when $\exists e \in E^-, \; h \cup B \models e$ i.e. when $fp(h)>0$.
A hypothesis $h$ is \emph{partially complete} when $\exists e \in E^+, \; h \cup B \models e$ i.e. when $tp(h)>0$ or $fn(h)<|E^+|$.
\section{Theoretical analysis}
\label{sec:proof}

In this section, we show the correctness of \name{}. Our proof follows the same outline as the proof of \citet{combo}.
We first assume that there is no constrain stage, i.e. that \name{} does not apply any noisy constraints.
With this assumption, we show that the generate, test, and combine stages return an optimal noisy solution. 
We then show that the constrain stage is optimally sound in that our noisy constraints never prune an optimal noisy solution from the hypothesis space.
We then use these results to prove the correctness of \name{} i.e. that \name{} returns an optimal noisy solution.

\subsection{Assumptions}
In the rest of this section, we assume a LFF input $(E, B, \mathcal{H}, C, cost_{B,E})$ where $\mathcal{H}_{C}$ is non-empty, i.e. we assume an optimal noisy solution always exists:
\begin{assumption}[\textbf{Existence of an optimal noisy solution}] \label{existencesol}Given a LFF input $(E, B, \mathcal{H}, C, cost_{B,E})$, we assume $\mathcal{H}_{C}$ is not empty.
\end{assumption}
\noindent
We follow LFF \cite{popper} and assume the background knowledge does not depend on any hypothesis:
\begin{assumption}[\textbf{Background knowledge independence}] \label{independentbk}We assume that no predicate symbol in the body of a rule in the background knowledge $B$ appears in the head of a rule in a hypothesis in $\mathcal{H}$. 
\end{assumption}

\subsection{Generate, Test, and Combine}
We recall the definition of a separable program \cite{combo}:
\begin{definition}
[\textbf{Separable program}]
A program $h$ is \emph{separable} when (i) it has at least two rules, and (ii) no predicate symbol in the head of a rule in $h$ also appears in the body of a rule in $h$.
\end{definition}
\noindent
In other words, a separable program cannot have recursive rules or invented predicate symbols.
\noindent
A program is \emph{non-separable} when it is not separable.

The correctness of the generate and test stages of \name{} follows from the correctness of the generate and test stages of \combo{} \cite{combo}:
\begin{lemma}
\label{lem_generate}
\name{} can generate and test every non-separable program.
\end{lemma}

\noindent
We now show that any optimal noisy solution is the union of non-separable partially complete programs:
\begin{lemma}\label{lem_opti_separable}
Let $I = (E, B, \mathcal{H}, C, cost_{B,E})$ be a LFF input, $h_o~\in~\mathcal{H}_{C}$ be an optimal noisy solution for $I$, and $S$ be the set of all non-separable partially complete programs for $I$.
Then $h_o = \bigcup_{i=1}^n h_i$ where each $h_i$ is in $S$.
\end{lemma}
\begin{proof}
An optimal noisy solution $h_o$ is either (a) non-separable or (b) separable.

For case (a), assume $h_o$ is non-separable and 
an optimal noisy solution for $I$.
If $h_o$ is partially complete then $h_o$ is in $S$.
If $h_o$ is not partially complete, then it is totally incomplete i.e. $fn(h_o)=|E^+|$.
We show $h_o$ is the empty hypothesis. By contradiction, we assume the opposite i.e. $size(h_o)>0$.
The empty hypothesis $[]$ has cost $cost_{B,E}([]) = |E^{+}|$.
\begin{align*}
cost_{B,E}(h_o) &= size(h_o)+fp(h_o)+fn(h_o)\\
&= size(h_o)+fp(h_o)+|E^+|\\
&> cost_{B,E}([])
\end{align*}
Then $h_o$ cannot be optimal, and $h_o$ is the empty hypothesis. In both cases $h_o = \bigcup_{i=1}^n h_i$ where each $h_i$ is in $S$.

For case (b), assume $h_o$ is separable and 
an optimal noisy solution for $I$.
\citet{combo} show (Lemma 2) that, since $h_o$ is separable, $h_o = \bigcup_{i=1}^n h_i$ where each $h_i$ is a non-separable program.
For contradiction, assume some non-empty $h_i$ is totally incomplete, i.e. $tp(h_i) = 0$. 
Let $h' = h_o \setminus h_i$. 
No predicate symbol in the body of $h'$ appears in the head of $h_i$. By Assumption \ref{independentbk}, no predicate symbol in the body of a rule in the background knowledge appears in the head of a rule in $h_i$. Therefore $h_i$ does not affect the logical consequences of $h'$. Since $tp(h_i) = 0$, then $tp(h_o) = tp(h')$ and $fn(h_o) = fn(h')$. Moreover, since $h_i$ is non-empty, then $size(h')<size(h_o)$. Since $h' \subseteq h_o$, then $fp(h') \leq fp(h_o)$.
\begin{align*}
cost_{B,E}(h_o) &= size(h_o)+fp(h_o)+fn(h_o)\\
&> size(h')+fp(h')+fn(h')\\
&> cost_{B,E}(h')\\
\end{align*}
Then $h_o$ cannot be optimal, and each $h_i$ is in $S$. 
Cases (a) and (b) are exhaustive so the proof is complete.
\end{proof}

\noindent
We show that the combine stage returns an optimal noisy solution:
\begin{lemma}
\label{lem_combine}
Let $I = (E, B, \mathcal{H}, C, cost_{B,E})$ be a LFF input and $S$ be the set of all non-separable partially complete programs for $I$. 
Then the combine stage returns an optimal noisy solution for $I$.
\end{lemma}
\begin{proof}
For contradiction, assume the opposite, which implies that the combine stage does not return any hypothesis, or returns a hypothesis $h$ that is a non-optimal noisy solution for $I$.
By Assumption \ref{existencesol}, an optimal noisy solution $h_o$ exists.
From Lemma \ref{lem_opti_separable}, this optimal noisy  solution verifies $h_o = \bigcup_{i=1}^n h_i$ where each $h_i \in S$. 
Then $h_o$ is a model for the combine encoding.
Since there are no recursive programs nor programs with invented predicates, the MaxSAT solver can reason about the coverage of a combination using the union of coverage of the individual programs in the combination.
The MaxSAT solver thus can reason about the cost of combinations as the sum of the sizes of the selected programs and the number of examples covered by a combination.
Moreover, there are finitely many programs in $S$.
Therefore, the MaxSAT solver finds a model that has minimal cost, which is a model with similar cost than $h_o$, and the MaxSAT solver returns an optimal noisy solution.
\end{proof}

\noindent
We show that \name{} (with only the generate, test, and combine stages) returns an optimal noisy solution:
\begin{lemma}
\label{lem_gen_one_opt}
\name{} returns an optimal noisy solution.
\end{lemma}
\begin{proof}
Lemma \ref{lem_generate} shows that \name{} can generate and test every non-separable program, of which there are finitely many.
Therefore, \name{} can generate and test every non-separable partially complete program.
Lemma \ref{lem_combine} shows that given a set of non-separable partially complete programs, the combine stage returns an optimal noisy solution.
Therefore \name{} returns an optimal noisy solution.
\end{proof}

\subsection{Constrain}
In this section, we show that the constrain stage never prunes an optimal noisy solution from the hypothesis space.  

\subsubsection{Generalisation constraints}
We show three results for generalisation constraints.
We recall that if a hypothesis $h_1$ is included in a hypothesis $h_2$, then (i) $size(h_1) \leq size(h_2)$ and (ii) $fp(h_1) \leq fp(h_2)$. We also recall that if a hypothesis $h_2$ is a generalisation of a hypothesis $h_1$, then (iii) $fp(h_1) \leq fp(h_2)$. 

We show our first result for generalisation constraints. If a hypothesis $h_1$ has $fn(h_1)$ false negatives, any of its generalisation can only reduce the number of false negatives by at most $fn(h_1)$. Therefore, given a hypothesis $h_1$, any hypothesis which includes a generalisation $h_2$ of $h_1$ verifying $size(h_2) > fn(h_1)+size(h_1)$ has greater MDL cost than $h_1$.
\begin{proposition}[Noisy generalisation constraint] \label{generalisationa} 
Let $h_1$ be a hypothesis, $h_2$ be a generalisation of $h_1$, and $size(h_2) > fn(h_1)+size(h_1)$.
Then $h_2$ cannot appear in an optimal noisy solution.
\end{proposition}
\begin{proof}
We assume (iv) $size(h_2) > fn(h_1)+size(h_1)$. We assume the opposite i.e. there exists an optimal noisy solution $h$ which includes $h_2$.
\begin{align*}
cost_{B,E}(h) &= size(h)+fp(h)+fn(h)\\ 
&\geq size(h_2)+fp(h_2)+fn(h) \text{ using (i) and (ii)}\\
&\geq size(h_2)+fp(h_1)+fn(h) \text{ using (iii)}\\
&> fn(h_1)+size(h_1)+fp(h_1)+fn(h)\\
&\quad \text{ using (iv)}\\
&\geq cost_{B,E}(h_1)
\end{align*}
Therefore the MDL cost of $h$ is greater than the cost of $h_1$ and $h$ is not an optimal noisy solution.
\end{proof}

\noindent
We show our second result for generalisation constraints. If a hypothesis $h_2$ is a generalisation of $h_1$, then $h_2$ has at least as many false positives as $h_1$, and therefore $h_2$ has a cost at least $fp(h_1)+size(h_2)$. We show the cost of any hypothesis containing $h_2$ is greater than the cost of the empty hypothesis when $size(h_2) > |E^+|-fp(h_1)$:
\begin{proposition}[Noisy generalisation constraint] \label{generalisationb} 
Let $h_1$ be a hypothesis, $h_2$ be a generalisation of $h_1$, and $size(h_2) > |E^+|-fp(h_1)$.
Then $h_2$ cannot appear in an optimal noisy solution.
\end{proposition}
\begin{proof}
We assume (iv) $size(h_2) > |E^+|-fp(h_1)$. We assume the opposite i.e. there exists an optimal noisy solution $h$ which includes $h_2$. The empty hypothesis, denoted as $[]$, has cost $cost_{B,E}([]) = |E^{+}|$.
\begin{align*}
cost_{B,E}(h) &= size(h)+fp(h)+fn(h)\\ 
&\geq size(h_2)+fp(h_2)+fn(h) \text{ using (i) and (ii)}\\
&\geq size(h_2)+fp(h_1)+fn(h) \text{ using (iii)}\\
&> |E^+|-fp(h_1)+fp(h_1)+fn(h)\text{ using (iv)}\\
&= |E^+|+fn(h)\\
&\geq cost_{B,E}([])
\end{align*}
Therefore the cost of $h$ is greater than the cost of the empty hypothesis and $h$ is not an optimal noisy solution.
\end{proof}

\noindent
We show our last result for generalisation constraints. If a hypothesis $h_2$ is a generalisation of $h_1$ and $size(h_2) > max_{mdl}-cost_{B,E}(h_1)+|E^{+}|+size(h_1)$ where $max_{mdl}$ is an upper bound on the MDL cost, then $h_2$ cannot appear in an optimal noisy solution:
\begin{proposition}[Noisy generalisation constraint] \label{generalisationc} 
Let $h_1$ be a hypothesis, $h_2$ be a generalisation of $h_1$, $h_o$ be an optimal noisy solution, $max_{mdl}$ be a cost such that $max_{mdl} \geq cost_{B,E}(h_o)$, and $size(h_2) > max_{mdl}-cost_{B,E}(h_1)+|E^{+}|+size(h_1)$.
Then $h_2$ cannot appear in an optimal noisy solution.
\end{proposition}
\begin{proof}
We assume (iv) $size(h_2) > max_{mdl}-cost_{B,E}(h_1)+|E^{+}|+size(h_1)$. We assume the opposite i.e. there exists an optimal noisy solution $h$ which includes $h_2$.
\begin{align*}
cost_{B,E}(h) &= size(h)+fp(h)+fn(h)\\ 
&\geq size(h_2)+fp(h_2)+fn(h) \text{ using (i) and (ii)}\\
&\geq size(h_2)+fp(h_1)+fn(h) \text{ using (iii)}\\
&> max_{mdl}-cost_{B,E}(h_1)+|E^{+}|+size(h_1)\\
&+fp(h_1)+fn(h) \text{ using (iv)}\\
&\geq max_{mdl}+|E^{+}|-fn(h_1)+fn(h)\\
&\geq max_{mdl} \text{ since } fn(h_1)-fn(h) \leq |E^{+}|\\
\end{align*}
Therefore the cost of $h$ is greater than $max_{mdl}$ and $h$ is not an optimal noisy solution.
\end{proof}
\noindent
In practice, in our implementation, we set $max_{mdl}$ to the cost of the current best hypothesis.

\subsubsection{Specialisation constraints}
We now show two results for specialisation constraints.
We recall that if a hypothesis $h$ verifies $h=h_2 \cup h_3$ where $h_2 \cap h_3 = \emptyset$, then (i) $size(h) = size(h_2)+size(h_3)$.

\noindent
We first show one intermediate result. 
This result provides a lower bound on the number of false negatives of a hypothesis $h=h_1 \cup h_2$ given the number of false negatives of its constituents $h_1$ and $h_2$:
\begin{lemma} Let $h_1$, $h_2$ and $h$ be three hypotheses such that $h=h_1 \cup h_2$. Then $fn(h)\geq fn(h_2)+fn(h_3)-|E^{+}|$.
\label{lemma_fn}
\end{lemma}
\begin{proof}
According to the inclusion–exclusion principle applied to finite sets: 
\begin{align*}
fn(h) &= fn(h_2)+fn(h_3)\\
&\quad-|\{e\in E^+ | B\cup h_2 \not\models e \text{ and } B\cup h_3 \not\models e\}|\\
fn(h)&\geq fn(h_2)+fn(h_3)-|E^{+}|
\end{align*}
\end{proof}


\noindent
We now show our first result for specialisation constraints. If a hypothesis $h_1$ has $tp(h_1)$ true positives, any of its specialisations can have at most $tp(h_1)$ true positives. Therefore, any hypothesis which includes a specialisation of $h_1$ with size greater than $tp(h_1)$ does not result in a positive compression of the positive examples and cannot appear in an optimal noisy solution:
\begin{proposition}[Noisy specialisation constraint] \label{specialisationa} 
Let $h_1$ be a hypothesis, $h_2$ be a specialisation of $h_1$, and $size(h_2)>tp(h_1)$.
Then $h_2$ cannot appear in an optimal noisy solution.
\end{proposition}
\begin{proof}
Since $h_2$ is a specialisation of a $h_1$, then (ii) $fn(h_1) \leq fn(h_2)$. Moreover, we know (iii) $|E^+|=fn(h_1)+tp(h_1)$. 
We assume (iv) $size(h_2) > tp(h_1)$. We assume the opposite i.e. there exists an optimal noisy solution $h$ which includes $h_2$. In other words, there exists a non-empty hypothesis $h_3$ such that $h=h_2 \cup h_3$ and $h_2 \cap h_3 = \emptyset$. Since $h_3$ is included in $h$, then (v) $fp(h_3) \leq fp(h)$.
\begin{align*}
cost_{B,E}(h) &= size(h)+fp(h)+fn(h)\\
&\geq size(h_2)+size(h_3)+fp(h_3)+fn(h)\\
&\quad \text{ using (i) and (v)}\\ 
&> tp(h_1)+size(h_3)+fp(h_3)+fn(h)\text{ using (iv)}\\ 
&\geq size(h_3)+fp(h_3)+fn(h)+|E^+|-fn(h_1)\\
&\quad \text{ using (iii)}\\
&\geq size(h_3)+fp(h_3)+fn(h)+|E^+|-fn(h_2)\\
&\quad \text{ using (ii)}\\
&\geq size(h_3)+fp(h_3)+fn(h_3)\text{ using Lemma \ref{lemma_fn}}\\
&= cost_{B,E}(h_3)
\end{align*}
Therefore the cost of $h$ is greater than the cost of $h_3$ and $h$ is not an optimal noisy solution.
\end{proof}

\noindent
We show our second result for specialisation constraints. If a hypothesis $h_1$ has $fp(h_1)$ false positives, any of its specialisations can only reduce the number of false positives by at most $fp(h_1)$. Therefore, any hypothesis which includes a specialisation $h_2$ of $h_1$ such that $size(h_2)> size(h_1)+fp(h_1)$ cannot appear in an optimal noisy solution:
\begin{proposition}[Noisy specialisation constraint] \label{specialisationb}
Let $h_1$ be a hypothesis, $h_2$ be a specialisation of $h_1$, and $size(h_2)> size(h_1)+fp(h_1)$.
Then $h_2$ cannot appear in an optimal noisy solution.
\end{proposition}
\begin{proof}
We assume (ii) $size(h_2)> size(h_1)+fp(h_1)$. We assume the opposite i.e. there exists an optimal noisy solution $h$ which includes $h_2$. In other words, there exists a non-empty hypothesis $h_3$ such that $h=h_2 \cup h_3$ and $h_2 \cap h_3 = \emptyset$. We show $h$ has greater cost than $h'=h_1 \cup h_3$. Since $h_2$ is a specialisation of $h_1$, then $h=h_2 \cup h_3$ is a specialisation of $h'=h_1 \cup h_3$ and (iii) $fn(h') \leq fn(h)$.
\begin{align*}
cost_{B,E}(h) &= size(h)+fp(h)+fn(h)\\
&= size(h_2)+size(h_3)+fp(h)+fn(h) \text{ using (i)}\\ 
&> size(h_1)+fp(h_1)+size(h_3)+fp(h)\\
&\quad +fn(h) \text{ using (ii)}\\ 
&\geq size(h')+fp(h_1)+fp(h)+fn(h)\\ 
&\geq size(h')+fp(h_1)+fp(h)+fn(h')\text{ using (iii)}\\ 
\end{align*}
We show $fp(h_1)+fp(h) \geq fp(h')$:
\begin{align*}
fp(h_1)+fp(h) &= |\{e\in E^- | B \cup h_1 \models e\}|\\ 
& \quad + \{e\in E^- | B \cup h_2 \models e \text{ or } B \cup h_3 \models e\}\\
&\geq|\{e\in E^- | B \cup h_1 \models e\}|\\
&\quad +\{e\in E^- | B \cup h_3 \models e\}\\
&\geq \{e\in E^- | B \cup h_1 \models e \text{ or } B \cup h_3 \models e\}\\
&\geq fp(h')
\end{align*}
Therefore $cost_{B,E}(h) > cost_{B,E}(h')$ and $h$ is not an optimal noisy solution.
\end{proof}

\subsection{\name{} Correctness}
We show the correctness of \name{}:
\begin{theorem}[Correctness]
\name{} returns an optimal noisy solution.
\end{theorem}
\begin{proof}
Lemma \ref{lem_gen_one_opt} shows that \name{} returns an optimal solution if there is no constrain stage. 
Moreover, noisy generalisation and specialisation constraints never prune an optimal noisy solution according to Propositions \ref{generalisationa} to \ref{specialisationb}. Therefore, \name{} returns an optimal noisy solution.
\end{proof}

\section{Experiments}
\label{sec:exp}

\subsection{Experimental domains}
We describe the characteristics of the domains and tasks used in our experiments in Tables \ref{tab:dataset} and \ref{tab:tasks}. Optimal solutions are unknown for \emph{alzheimer} and \emph{wn18rr} tasks. Figure \ref{fig:sols} shows example solutions for some of the tasks.

\begin{table}[ht!]
\footnotesize
\centering
\begin{tabular}{@{}l|cccc@{}}
\textbf{Task} & \textbf{\# examples} & \textbf{\# relations} & \textbf{\# constants} & \textbf{\# facts}\\
\midrule
\emph{md} & 54 & 12 & 13 & 29 \\
\emph{buttons} & 530 & 13 & 60 & 656 \\
\emph{coins} & 2544 & 9 & 110 & 1101 \\
\emph{rps} & 464 & 6 & 64 & 405 \\\midrule
\emph{zendo1} & 100 & 16 & 1049 & 2270\\
\emph{zendo2} & 100 & 16 & 1047 & 2184\\
\emph{zendo3} & 100 & 16 & 1100 & 2320\\
\emph{zendo4} & 100 & 16 & 987 & 2087\\\midrule
\emph{dropk} & 200 & 10 & $\infty$ & $\infty$\\
\emph{evens} & 200 & 10 & $\infty$ & $\infty$\\
\emph{reverse} & 200 & 10 & $\infty$ & $\infty$\\
\emph{sorted} & 200 & 10 & $\infty$ & $\infty$\\\midrule
\emph{alz-acetyl} & 1326 & 32 & 149 & 628\\
\emph{alz-amine} & 686 & 32 & 149 & 628\\
\emph{alz-mem} & 642 & 32 & 149 & 628\\
\emph{alz-toxic} & 886 & 32 & 149 & 628\\
\midrule
\emph{wn18rr1} & 1846 & 11 & 40559 & 86835\\
\emph{wn18rr2} & 2276 & 11 & 40559 & 86835\\
\end{tabular}
\caption{
Experimental domain description.
}

\label{tab:dataset}
\end{table}

\begin{table}[ht!]
\footnotesize
\centering
\begin{tabular}{@{}l|cccc@{}}
\textbf{Task} & \textbf{\#rules} & \textbf{\#literals} & \textbf{max rule size} & \textbf{recursion}\\
\midrule
\emph{md} & 2 & 11 & 6 & no\\
\emph{buttons} & 10 & 61 & 7& no\\
\emph{coins} & 16 & 45 & 7& no\\
\emph{rps} & 4 & 25 & 7& no\\
\midrule
\emph{zendo1} & 1 & 7 & 7 & no\\
\emph{zendo2} & 2 & 14 & 7 & no\\
\emph{zendo3} & 3 & 20 & 7& no\\
\emph{zendo4} & 4 & 23 & 7& no\\
\midrule
\emph{dropk} & 2 & 7 & 4& yes\\
\emph{evens} & 2 & 7 & 5& yes\\
\emph{reverse} & 2 & 8 & 5& yes\\
\emph{sorted} & 2 & 9 & 6& yes\\
\end{tabular}
\caption{
Statistics about the optimal solutions for each task.
For instance, the optimal solution for \emph{buttons} has 10 rules and 61 literals and the largest rule has 7 literals.
}

\label{tab:tasks}
\end{table}


\paragraph{IGGP.}
In \emph{inductive general game playing} (IGGP)  \cite{iggp} the task is to induce a hypothesis to explain game traces from the general game playing competition \cite{ggp}.
IGGP is notoriously difficult for machine learning approaches.
The currently best-performing system can only learn perfect solutions for 40\% of the tasks.
Moreover, although seemingly a toy problem, IGGP is representative of many real-world problems, such as inducing semantics of programming languages \cite{DBLP:conf/ilp/BarthaC19}. 
We use four games: \emph{minimal decay (md)}, \emph{buttons}, \emph{rock - paper - scissors (rps)}, and \emph{coins}.

\paragraph{Zendo.} Zendo is an inductive game in which one player, the Master, creates a rule for structures made of pieces with varying attributes to follow. The other players, the Students, try to discover the rule by building and studying structures which are labelled by the Master as following or breaking the rule. The first student to correctly state the rule wins. We learn four increasingly complex rules for structures made of at most 5 pieces of varying color, size, orientation and position. 
Zendo is a challenging game that has attracted much interest in cognitive science \cite{zendo}.



\paragraph{Program Synthesis.} This dataset includes list transformation tasks. It involves learning recursive programs which have been identified as a difficult challenge for ILP systems \cite{ilp20}. 

\paragraph{Alzheimer.} The goal is to learn rules describing four important biochemical properties of drugs: low toxicity, high acetyl cholinesterase inhibition, high reversal of scopolamine-induced memory impairment, and high inhibition of amine re-uptake \cite{alzheimer}. The background knowledge contains information about the physical and chemical properties of substituents such as their hydrophobicity and polarity and the relations between various physical and chemical constants. Positive and negative examples are pairwise comparisons of drugs.

\paragraph{Wn18rr.} Wn18rr \cite{wn18rr} is a link prediction dataset created from wn18, which is a subset of WordNet. 
WordNet is designed to produce a dictionary, and support automatic text analysis. Entities are  word senses, and relations define lexical relations between entities. 
We predict the relations \emph{memberofdomainregion} and \emph{verbgroup}.

\subsection{Experimental Setup}
We measure the mean and standard error of the predictive accuracy and learning time.
We use a 3.8 GHz 8-Core Intel Core i7 with 32GB of ram.
The systems use a single CPU.

\subsection{Experimental Methods}

To answer \textbf{Q1} and \textbf{Q2} we compare \name{} against \noisypopper{}, \popper{} and \ale{} which we describe below.

\begin{description}
\item[\popper{} and \noisypopper{}] \popper{} \cite{popper} and \noisypopper{} \cite{noisypopper} use identical biases to \name{} so the comparison is direct, i.e. fair.
We use \popper{} 2.0.0\footnote{\url{https://github.com/logic-and-learning-lab/Popper/releases/tag/v2.0.0}}.

\item[\ale{}] \ale{} \cite{aleph} excels at learning many large non-recursive rules and should excel at the \emph{trains} and \emph{iggp} tasks.
Although \ale{} can learn recursive programs, it struggles to do so.
\name{} and \ale{} use similar biases so the comparison can be considered reasonably fair.

\end{description}
\noindent
We also considered other ILP systems.
We considered \ilasp{} \cite{ilasp}.
However, \ilasp{} builds on \aspal{} \cite{aspal} and first precomputes every possible rule in a hypothesis space, which is infeasible for our datasets.
In addition, \ilasp{} cannot learn Prolog programs so is unusable in the synthesis tasks.
For instance, it would require precomputing $10^{15}$ rules for the \emph{iggp-coins} task.
We also considered \textsc{Metagol} \cite{metagol}, \textsc{Hexmil} \cite{hexmil} and \textsc{Louise} \cite{louise}, which are metarule-based approaches.

\begin{figure*}
\centering
\footnotesize
\begin{lstlisting}[caption=minimal-decay \label{md}]
next_value(A,B):- c5(B),pressButton(D),player(C),does(A,C,D).
next_value(A,B):- player(D),true_value(A,C),noop(E),does(A,D,E),my_succ(B,C).
\end{lstlisting}

\begin{lstlisting}[caption=buttons \label{buttons}]
next(A,B):-c_p(B),c_c(C),does(A,D,C),my_true(A,B),my_input(D,C).
next(A,B):-my_input(C,E),c_p(D),my_true(A,D),c_b(E),does(A,C,E),c_q(B).
next(A,B):-my_input(C,D),not_my_true(A,B),does(A,C,D),c_p(B),c_a(D).
next(A,B):-c_a(C),does(A,D,C),my_true(A,B),c_q(B),my_input(D,C).
next(A,B):-my_input(C,E),c_p(B),my_true(A,D),c_b(E),does(A,C,E),c_q(D).
next(A,B):-c_c(D),my_true(A,C),c_r(B),role(E),does(A,E,D),c_q(C).
next(A,B):-my_true(A,C),my_succ(C,B).
next(A,B):-my_input(C,D),does(A,C,D),my_true(A,B),c_r(B),c_b(D).
next(A,B):-my_input(C,D),does(A,C,D),my_true(A,B),c_r(B),c_a(D).
next(A,B):-my_true(A,E),c_c(C),does(A,D,C),c_q(B),c_r(E),my_input(D,C).
\end{lstlisting}

\begin{lstlisting}[caption=rps\label{rps}]
next_score(A,B,C):-does(A,B,E),different(G,B),my_true_score(A,B,F),beats(E,D),my_succ(F,C),does(A,G,D).
next_score(A,B,C):-different(G,B),beats(D,F),my_true_score(A,E,C),does(A,G,D),does(A,E,F).
next_score(A,B,C):-my_true_score(A,B,C),does(A,B,D),does(A,E,D),different(E,B).
\end{lstlisting}

\begin{lstlisting}[caption=coins\label{coins}]
next_cell(A,B,C):-does_jump(A,E,F,D),role(E),different(B,D),my_true_cell(A,B,C),different(F,B).
next_cell(A,B,C):-my_pos(E),role(D),c_zerocoins(C),does_jump(A,D,B,E).
next_cell(A,B,C):-role(D),does_jump(A,D,E,B),c_twocoins(C),different(B,E).
next_cell(A,B,C):-does_jump(A,F,E,D),role(F),my_succ(E,B),my_true_cell(A,B,C),different(E,D).
\end{lstlisting}

\centering
\begin{lstlisting}[caption=dropk]
dropk(A,B,C):- tail(A,C),one(B).
dropk(A,B,C):- decrement(B,E),tail(A,D),dropk(D,E,C).
\end{lstlisting}

\centering
\begin{lstlisting}[caption=evens]
evens(A):- empty(A).
evens(A):- head(A,B),tail(A,C),even(B),evens(C).
\end{lstlisting}

\centering
\begin{lstlisting}[caption=reverse]
reverse(A,B):- empty_out(B),empty(A).
reverse(A,B):- head(A,D),tail(A,E),reverse(E,C),append(C,D,B).
\end{lstlisting}

\centering
\begin{lstlisting}[caption=sorted]
sorted(A):-tail(A,B),empty(B).
sorted(A):-tail(A,D),head(A,B),head(D,C),geq(C,B),sorted(D).
\end{lstlisting}

\begin{lstlisting}[caption=zendo1\label{zendo1}]
zendo1(A):- piece(A,C),size(C,B),blue(C),small(B),contact(C,D),red(D).
\end{lstlisting}

\centering
\begin{lstlisting}[caption=zendo2\label{zendo2}]
zendo2(A):- piece(A,B),piece(A,D),piece(A,C),green(D),red(B),blue(C).
zendo2(A):- piece(A,D),piece(A,B),coord1(B,C),green(D),lhs(B),coord1(D,C).
\end{lstlisting}
\centering

\begin{lstlisting}[caption=zendo3\label{zendo3}]
zendo3(A):- piece(A,D),blue(D),coord1(D,B),piece(A,C),coord1(C,B),red(C).
zendo3(A):- piece(A,D),contact(D,C),rhs(D),size(C,B),large(B).
zendo3(A):- piece(A,B),upright(B),contact(B,D),blue(D),size(D,C),large(C).
\end{lstlisting}
\centering

\begin{lstlisting}[caption=zendo4\label{zendo4}]
zendo4(A):- piece(A,C),contact(C,B),strange(B),upright(C).
zendo4(A):- piece(A,D),contact(D,C),coord2(C,B),coord2(D,B).
zendo4(A):- piece(A,D),contact(D,C),size(C,B),red(D),medium(B).
zendo4(A):- piece(A,D),blue(D),lhs(D),piece(A,C),size(C,B),small(B).\end{lstlisting}

\caption{Example solutions.}
\label{fig:sols}
\end{figure*}

\section{Example \name{} Output}
\label{sec:outputs}
Figure \ref{fig:rps} shows an example \name{} output on the \emph{iggp-rps (10)} task.

\begin{figure*}[ht!]
\begin{lstlisting}
********************
New best hypothesis:
tp:76 fn:63 tn:285 fp:40 size:2 cost:105
next_score(A,B,C):- my_true_score(A,B,C).
********************
new best cost 105
Searching programs of size: 2
Searching programs of size: 3
Searching programs of size: 4
Searching programs of size: 5
********************
New best hypothesis:
tp:74 fn:65 tn:291 fp:34 size:5 cost:104
next_score(A,B,C):- my_true_score(A,D,C),does(A,D,E),my_true_score(A,B,C),does(A,B,E).
********************
new best cost 104
Searching programs of size: 6
Searching programs of size: 7
********************
New best hypothesis:
tp:71 fn:68 tn:323 fp:2 size:11 cost:81
next_score(A,B,C):- does(A,D,E),does(A,B,E),my_true_score(A,B,C),different(D,B).
next_score(A,B,C):- my_true_score(A,G,C),beats(D,E),different(G,F),does(A,F,D),does(A,B,E).
********************
new best cost 81
********************
New best hypothesis:
tp:103 fn:36 tn:320 fp:5 size:18 cost:59
next_score(A,B,C):- does(A,D,E),does(A,B,E),my_true_score(A,B,C),different(D,B).
next_score(A,B,C):- my_true_score(A,G,C),beats(D,E),different(G,F),does(A,F,D),does(A,B,E).
next_score(A,B,C):- my_true_score(A,B,E),beats(D,G),does(A,F,G),player(F),does(A,B,D),my_succ(E,C).
********************
********** SOLUTION **********
Precision:0.95 Recall:0.74 TP:103 FN:36 TN:320 FP:5 Size:18 cost:59
next_score(A,B,C):- does(A,D,E),does(A,B,E),my_true_score(A,B,C),different(D,B).
next_score(A,B,C):- my_true_score(A,G,C),beats(D,E),different(G,F),does(A,F,D),does(A,B,E).
next_score(A,B,C):- my_true_score(A,B,E),beats(D,G),does(A,F,G),player(F),does(A,B,D),my_succ(E,C).
******************************
Num. programs: 8820
Generate:
	Called: 8821 times 	 Total: 21.51 	 Mean: 0.002 	 Max: 1.930 	 Percentage: 34%
Test:
	Called: 8820 times 	 Total: 18.39 	 Mean: 0.002 	 Max: 0.076 	 Percentage: 29%
Constrain:
	Called: 8820 times 	 Total: 12.81 	 Mean: 0.001 	 Max: 0.013 	 Percentage: 20%
Combine:
	Called: 5 times 	 Total: 0.18 	 Mean: 0.035 	 Max: 0.056 	 Percentage: 0%
Total operation time: 61.85s
Total execution time: 64.03s
\end{lstlisting}
\caption{Example \name{} output on the \emph{iggp-rps (10)} task.}
\label{fig:rps}
\end{figure*}

\section{Experimental results}\label{sec:results}

\subsection{Comparison against SOTA}
Tables \ref{tab:time_appendix} and \ref{tab:accuracies_appendix} show the learning times and the predictive accuracies with a timeout of 1200s for \name{}, \popper{}, \ale{} and \noisypopper{} for the tasks tested.


\begin{table*}[ht]
\centering
\footnotesize
\begin{tabular}{@{}l|cccc@{}}
\textbf{Task} & \textbf{\name} & \textbf{\popper} & \textbf{\ale} & \textbf{\noisypopper}\\
\midrule
\emph{iggp-md (0)} & \textbf{1 $\pm$ 0} & 11 $\pm$ 0 & 3 $\pm$ 0 & \emph{timeout}\\
\emph{iggp-md (10)} & \textbf{2 $\pm$ 0} & 11 $\pm$ 1 & 17 $\pm$ 2 & \emph{timeout}\\
\emph{iggp-md (20)} & \textbf{2 $\pm$ 0} & 16 $\pm$ 1 & 29 $\pm$ 3 & \emph{timeout}\\
\emph{iggp-buttons (0)} & \textbf{7 $\pm$ 0} & \textbf{7 $\pm$ 0} & 543 $\pm$ 8 & \emph{timeout}\\
\emph{iggp-buttons (10)} & \textbf{9 $\pm$ 0} & 12 $\pm$ 0 & \emph{timeout} & \emph{timeout}\\
\emph{iggp-buttons (20)} & \textbf{10 $\pm$ 0} & 12 $\pm$ 0 & \emph{timeout} & \emph{timeout}\\
\emph{iggp-coins (0)} & \textbf{138 $\pm$ 1} & 147 $\pm$ 2 & \emph{timeout} & \emph{timeout}\\
\emph{iggp-coins (10)} & \textbf{141 $\pm$ 2} & 146 $\pm$ 2 & \emph{timeout} & \emph{timeout}\\
\emph{iggp-coins (20)} & \textbf{144 $\pm$ 2} & 148 $\pm$ 1 & \emph{timeout} & \emph{timeout}\\
\emph{iggp-rps (0)} & 50 $\pm$ 1 & 53 $\pm$ 1 & \textbf{26 $\pm$ 2} & \emph{timeout}\\
\emph{iggp-rps (10)} & \textbf{60 $\pm$ 2} & 72 $\pm$ 2 & 884 $\pm$ 42 & \emph{timeout}\\
\emph{iggp-rps (20)} & \textbf{66 $\pm$ 1} & 75 $\pm$ 2 & \emph{timeout} & \emph{timeout}\\
\midrule
\emph{zendo1 (0)} & \textbf{13 $\pm$ 3} & 23 $\pm$ 4 & 14 $\pm$ 3 & \emph{timeout}\\
\emph{zendo1 (10)} & \textbf{77 $\pm$ 4} & 123 $\pm$ 2 & 130 $\pm$ 14 & \emph{timeout}\\
\emph{zendo1 (20)} & \textbf{100 $\pm$ 13} & 117 $\pm$ 4 & 192 $\pm$ 13 & \emph{timeout}\\
\emph{zendo2 (0)} & 48 $\pm$ 1 & 102 $\pm$ 2 & \textbf{14 $\pm$ 2} & \emph{timeout}\\
\emph{zendo2 (10)} & \textbf{48 $\pm$ 3} & 104 $\pm$ 2 & 244 $\pm$ 13 & \emph{timeout}\\
\emph{zendo2 (20)} & 118 $\pm$ 36 & \textbf{104 $\pm$ 3} & 258 $\pm$ 13 & \emph{timeout}\\
\emph{zendo3 (0)} & 49 $\pm$ 3 & 85 $\pm$ 3 & \textbf{15 $\pm$ 2} & \emph{timeout}\\
\emph{zendo3 (10)} & \textbf{62 $\pm$ 2} & 102 $\pm$ 2 & 151 $\pm$ 11 & \emph{timeout}\\
\emph{zendo3 (20)} & 190 $\pm$ 112 & \textbf{107 $\pm$ 2} & 171 $\pm$ 7 & \emph{timeout}\\
\emph{zendo4 (0)} & 43 $\pm$ 3 & 79 $\pm$ 4 & \textbf{14 $\pm$ 1} & \emph{timeout}\\
\emph{zendo4 (10)} & \textbf{58 $\pm$ 2} & 102 $\pm$ 9 & 103 $\pm$ 7 & \emph{timeout}\\
\emph{zendo4 (20)} & 122 $\pm$ 32 & \textbf{112 $\pm$ 4} & 134 $\pm$ 7 & \emph{timeout}\\
\midrule
\emph{dropk (0)} & 7 $\pm$ 1 & \textbf{4 $\pm$ 1} & 899 $\pm$ 161 & \emph{timeout}\\
\emph{dropk (10)} & \emph{timeout} & \emph{timeout} & \textbf{403 $\pm$ 175} & \emph{timeout}\\
\emph{dropk (20)} & \emph{timeout} & \emph{timeout} & \textbf{381 $\pm$ 179} & \emph{timeout}\\
\emph{evens (0)} & 7 $\pm$ 0 & 8 $\pm$ 0 & 22 $\pm$ 1 & \textbf{12 $\pm$ 1}\\
\emph{evens (10)} & \emph{timeout} & \emph{timeout} & \textbf{21 $\pm$ 3} & \emph{timeout}\\
\emph{evens (20)} & \emph{timeout} & \emph{timeout} & \textbf{22 $\pm$ 2} & \emph{timeout}\\
\emph{reverse (0)} & \textbf{45 $\pm$ 7} & 65 $\pm$ 9 & 1116 $\pm$ 84 & \emph{timeout}\\
\emph{reverse (10)} & \emph{timeout} & \emph{timeout} & \emph{timeout} & \emph{timeout}\\
\emph{reverse (20)} & \emph{timeout} & \emph{timeout} & \textbf{1166 $\pm$ 34} & \emph{timeout}\\
\emph{sorted (0)} & 31 $\pm$ 3 & \textbf{19 $\pm$ 2} & 486 $\pm$ 194 & \emph{timeout}\\
\emph{sorted (10)} & \emph{timeout} & \emph{timeout} & \textbf{134 $\pm$ 119} & \emph{timeout}\\
\emph{sorted (20)} & \emph{timeout} & \emph{timeout} & \textbf{139 $\pm$ 118} & \emph{timeout}\\
\midrule
\emph{alzheimer\_acetyl} & \textbf{133 $\pm$ 5} & 529 $\pm$ 28 & \emph{timeout} & \emph{timeout}\\
\emph{alzheimer\_amine} & \textbf{73 $\pm$ 3} & 152 $\pm$ 6 & 615 $\pm$ 24 & \emph{timeout}\\
\emph{alzheimer\_mem} & \textbf{79 $\pm$ 3} & 518 $\pm$ 49 & 773 $\pm$ 20 & \emph{timeout}\\
\emph{alzheimer\_toxic} & \textbf{61 $\pm$ 6} & 120 $\pm$ 22 & 511 $\pm$ 16 & \emph{timeout}\\
\midrule
\emph{wn18rr1} & 534 $\pm$ 14 & \textbf{499 $\pm$ 18} & \emph{timeout} & \emph{timeout}\\
\emph{wn18rr2} & \emph{timeout} & \emph{timeout} & \emph{timeout} & \emph{timeout}\\
\end{tabular}
\caption{
Learning times with a timeout of 1200s. We round times over one second to the nearest second. The error is standard error. Numbers in parenthesis indicate the level of noise added. The level of noise is unknown when unspecified.}
\label{tab:time_appendix}
\end{table*}

\begin{table*}[ht]
\centering
\footnotesize
\begin{tabular}{@{}l|cccc@{}}
\textbf{Task} & \textbf{\name} & \textbf{\popper} & \textbf{\ale} & \textbf{\noisypopper}\\
\midrule
\emph{iggp-md (0)} & 75 $\pm$ 0 & \textbf{100 $\pm$ 0} & \textbf{100 $\pm$ 0} & 75 $\pm$ 0\\
\emph{iggp-md (10)} & 65 $\pm$ 4 & 76 $\pm$ 6 & \textbf{85 $\pm$ 6} & 75 $\pm$ 0\\
\emph{iggp-md (20)} & 58 $\pm$ 4 & 68 $\pm$ 6 & \textbf{73 $\pm$ 7} & 70 $\pm$ 3\\
\emph{iggp-buttons (0)} & 80 $\pm$ 0 & \textbf{100 $\pm$ 0} & \textbf{100 $\pm$ 0} & 74 $\pm$ 0\\
\emph{iggp-buttons (10)} & \textbf{79 $\pm$ 1} & \textbf{79 $\pm$ 3} & 50 $\pm$ 0 & 74 $\pm$ 0\\
\emph{iggp-buttons (20)} & \textbf{77 $\pm$ 1} & 63 $\pm$ 1 & 50 $\pm$ 0 & 75 $\pm$ 1\\
\emph{iggp-coins (0)} & \textbf{100 $\pm$ 0} & \textbf{100 $\pm$ 0} & 50 $\pm$ 0 & 85 $\pm$ 0\\
\emph{iggp-coins (10)} & \textbf{100 $\pm$ 0} & 54 $\pm$ 1 & 50 $\pm$ 0 & 85 $\pm$ 0\\
\emph{iggp-coins (20)} & \textbf{100 $\pm$ 0} & 50 $\pm$ 0 & 50 $\pm$ 0 & 85 $\pm$ 0\\
\emph{iggp-rps (0)} & \textbf{100 $\pm$ 0} & \textbf{100 $\pm$ 0} & \textbf{100 $\pm$ 0} & 61 $\pm$ 3\\
\emph{iggp-rps (10)} & \textbf{100 $\pm$ 0} & 63 $\pm$ 2 & 73 $\pm$ 3 & 63 $\pm$ 3\\
\emph{iggp-rps (20)} & \textbf{100 $\pm$ 0} & 59 $\pm$ 2 & 50 $\pm$ 0 & 64 $\pm$ 3\\
\midrule
\emph{zendo1 (0)} & 99 $\pm$ 0 & \textbf{100 $\pm$ 0} & \textbf{100 $\pm$ 0} & 98 $\pm$ 0\\
\emph{zendo1 (10)} & \textbf{99 $\pm$ 0} & 82 $\pm$ 1 & 82 $\pm$ 1 & 98 $\pm$ 0\\
\emph{zendo1 (20)} & \textbf{99 $\pm$ 0} & 74 $\pm$ 1 & 76 $\pm$ 1 & \textbf{99 $\pm$ 0}\\
\emph{zendo2 (0)} & \textbf{100 $\pm$ 0} & \textbf{100 $\pm$ 0} & \textbf{100 $\pm$ 0} & 82 $\pm$ 1\\
\emph{zendo2 (10)} & \textbf{100 $\pm$ 0} & 64 $\pm$ 1 & 65 $\pm$ 1 & 82 $\pm$ 0\\
\emph{zendo2 (20)} & \textbf{100 $\pm$ 0} & 58 $\pm$ 1 & 60 $\pm$ 1 & 82 $\pm$ 0\\
\emph{zendo3 (0)} & 98 $\pm$ 0 & \textbf{99 $\pm$ 0} & \textbf{99 $\pm$ 0} & 82 $\pm$ 0\\
\emph{zendo3 (10)} & \textbf{98 $\pm$ 0} & 72 $\pm$ 1 & 72 $\pm$ 1 & 82 $\pm$ 0\\
\emph{zendo3 (20)} & \textbf{97 $\pm$ 1} & 68 $\pm$ 1 & 70 $\pm$ 1 & 81 $\pm$ 0\\
\emph{zendo4 (0)} & 98 $\pm$ 0 & \textbf{99 $\pm$ 0} & \textbf{99 $\pm$ 0} & 87 $\pm$ 0\\
\emph{zendo4 (10)} & \textbf{96 $\pm$ 0} & 81 $\pm$ 1 & 82 $\pm$ 1 & 87 $\pm$ 0\\
\emph{zendo4 (20)} & \textbf{94 $\pm$ 1} & 74 $\pm$ 1 & 77 $\pm$ 1 & 88 $\pm$ 0\\
\midrule
\emph{dropk (0)} & \textbf{100 $\pm$ 0} & \textbf{100 $\pm$ 0} & 55 $\pm$ 5 & 53 $\pm$ 0\\
\emph{dropk (10)} & \textbf{100 $\pm$ 0} & 54 $\pm$ 1 & 50 $\pm$ 0 & 55 $\pm$ 2\\
\emph{dropk (20)} & \textbf{100 $\pm$ 0} & 54 $\pm$ 1 & 50 $\pm$ 0 & 55 $\pm$ 2\\
\emph{evens (0)} & \textbf{100 $\pm$ 0} & \textbf{100 $\pm$ 0} & 57 $\pm$ 3 & 86 $\pm$ 0\\
\emph{evens (10)} & \textbf{100 $\pm$ 0} & 52 $\pm$ 1 & 52 $\pm$ 1 & \textbf{100 $\pm$ 0}\\
\emph{evens (20)} & \textbf{100 $\pm$ 0} & 51 $\pm$ 0 & 51 $\pm$ 0 & \textbf{100 $\pm$ 0}\\
\emph{reverse (0)} & \textbf{100 $\pm$ 0} & \textbf{100 $\pm$ 0} & 50 $\pm$ 0 & 52 $\pm$ 0\\
\emph{reverse (10)} & \textbf{100 $\pm$ 0} & 52 $\pm$ 0 & 50 $\pm$ 0 & 52 $\pm$ 0\\
\emph{reverse (20)} & \textbf{100 $\pm$ 0} & 52 $\pm$ 0 & 50 $\pm$ 0 & 51 $\pm$ 0\\
\emph{sorted (0)} & \textbf{100 $\pm$ 0} & \textbf{100 $\pm$ 0} & 68 $\pm$ 5 & 76 $\pm$ 1\\
\emph{sorted (10)} & \textbf{100 $\pm$ 0} & 64 $\pm$ 2 & 63 $\pm$ 2 & 77 $\pm$ 1\\
\emph{sorted (20)} & \textbf{100 $\pm$ 0} & 58 $\pm$ 1 & 56 $\pm$ 2 & 75 $\pm$ 1\\
\midrule
\emph{alzheimer\_acetyl} & \textbf{68 $\pm$ 1} & 56 $\pm$ 0 & 50 $\pm$ 0 & 61 $\pm$ 1\\
\emph{alzheimer\_amine} & \textbf{76 $\pm$ 1} & 69 $\pm$ 1 & 73 $\pm$ 1 & 56 $\pm$ 2\\
\emph{alzheimer\_mem} & \textbf{63 $\pm$ 1} & 51 $\pm$ 0 & 61 $\pm$ 1 & 62 $\pm$ 1\\
\emph{alzheimer\_toxic} & 74 $\pm$ 1 & 64 $\pm$ 1 & \textbf{83 $\pm$ 1} & 68 $\pm$ 1\\
\midrule
\emph{wn18rr1} & \textbf{98 $\pm$ 0} & 95 $\pm$ 1 & 50 $\pm$ 0 & 83 $\pm$ 0\\
\emph{wn18rr2} & \textbf{79 $\pm$ 1} & 78 $\pm$ 1 & 50 $\pm$ 0 & 62 $\pm$ 1\\
\end{tabular}
\caption{
Predictive accuracies. We round accuracies to integer values. The error is standard error. Numbers in parentheses indicate the level of noise added. The level of noise is unknown when unspecified.
}
\label{tab:accuracies_appendix}
\end{table*}

\subsection{Impact of the noisy constraints}
We compare \name{} with and without noisy constraints. 
Table \ref{tab:constraints_time_appendix} shows the learning times for \name{} with and without noisy constraints. Table \ref{tab:constraints_acc_appendix} shows the predictive accuracies for \name{} with and without noisy constraints. It shows the predictive accuracies are equal or higher
with noisy constraints. As shown by Propositions \ref{generalisationa} to \ref{specialisationb}, the constraints imposed by \name{} are optimally sound, i.e. they do not prune optimal noisy solutions. Therefore, \name{} can learn an optimal noisy solution with and without noisy constraints. However, \name{} without constraints searches a larger space and may not find a good solution within the time limit. For instance, \name{} without constraints reaches timeout for all \emph{alzheimer} tasks.

\begin{table}[ht]
\centering
\footnotesize
\begin{tabular}{@{}l|ccc@{}}
\textbf{Task} & \textbf{Without} & \textbf{With} & \textbf{Difference}\\
\midrule
\emph{iggp-md (0)} & 75 $\pm$ 0 & 75 $\pm$ 0& 0\%\\
\emph{iggp-md (10)} & 60 $\pm$ 4 & 65 $\pm$ 4& \textbf{+8\%}\\
\emph{iggp-md (20)} & 52 $\pm$ 2 & 58 $\pm$ 4& \textbf{+11\%}\\
\emph{iggp-buttons (0)} & 80 $\pm$ 0 & 80 $\pm$ 0& 0\%\\
\emph{iggp-buttons (10)} & 78 $\pm$ 1 & 79 $\pm$ 1& \textbf{+1\%}\\
\emph{iggp-buttons (20)} & 78 $\pm$ 1 & 77 $\pm$ 1& -1\%\\
\emph{iggp-coins (0)} & 100 $\pm$ 0 & 100 $\pm$ 0& 0\%\\
\emph{iggp-coins (10)} & 100 $\pm$ 0 & 100 $\pm$ 0& 0\%\\
\emph{iggp-coins (20)} & 100 $\pm$ 0 & 100 $\pm$ 0& 0\%\\
\emph{iggp-rps (0)} & 100 $\pm$ 0 & 100 $\pm$ 0& 0\%\\
\emph{iggp-rps (10)} & 100 $\pm$ 0 & 100 $\pm$ 0& 0\%\\
\emph{iggp-rps (20)} & 100 $\pm$ 0 & 100 $\pm$ 0& 0\%\\
\midrule
\emph{zendo1 (0)} & 99 $\pm$ 0 & 99 $\pm$ 0& 0\%\\
\emph{zendo1 (10)} & 99 $\pm$ 0 & 99 $\pm$ 0& 0\%\\
\emph{zendo1 (20)} & 98 $\pm$ 0 & 99 $\pm$ 0& \textbf{+1\%}\\
\emph{zendo2 (0)} & 100 $\pm$ 0 & 100 $\pm$ 0& 0\%\\
\emph{zendo2 (10)} & 100 $\pm$ 0 & 100 $\pm$ 0& 0\%\\
\emph{zendo2 (20)} & 97 $\pm$ 2 & 100 $\pm$ 0& \textbf{+3\%}\\
\emph{zendo3 (0)} & 98 $\pm$ 0 & 98 $\pm$ 0& 0\%\\
\emph{zendo3 (10)} & 98 $\pm$ 0 & 98 $\pm$ 0& 0\%\\
\emph{zendo3 (20)} & 97 $\pm$ 0 & 97 $\pm$ 1& 0\%\\
\emph{zendo4 (0)} & 98 $\pm$ 0 & 98 $\pm$ 0& 0\%\\
\emph{zendo4 (10)} & 96 $\pm$ 1 & 96 $\pm$ 0& 0\%\\
\emph{zendo4 (20)} & 96 $\pm$ 1 & 94 $\pm$ 1& -2\%\\
\midrule
\emph{dropk (0)} & 100 $\pm$ 0 & 100 $\pm$ 0& 0\%\\
\emph{dropk (10)} & 100 $\pm$ 0 & 100 $\pm$ 0& 0\%\\
\emph{dropk (20)} & 100 $\pm$ 0 & 100 $\pm$ 0& 0\%\\
\emph{evens (0)} & 100 $\pm$ 0 & 100 $\pm$ 0& 0\%\\
\emph{evens (10)} & 100 $\pm$ 0 & 100 $\pm$ 0& 0\%\\
\emph{evens (20)} & 100 $\pm$ 0 & 100 $\pm$ 0& 0\%\\
\emph{reverse (0)} & 51 $\pm$ 0 & 100 $\pm$ 0& \textbf{+96\%}\\
\emph{reverse (10)} & 51 $\pm$ 0 & 100 $\pm$ 0& \textbf{+96\%}\\
\emph{reverse (20)} & 50 $\pm$ 0 & 100 $\pm$ 0& \textbf{+100\%}\\
\emph{sorted (0)} & 90 $\pm$ 3 & 100 $\pm$ 0& \textbf{+11\%}\\
\emph{sorted (10)} & 88 $\pm$ 3 & 100 $\pm$ 0& \textbf{+13\%}\\
\emph{sorted (20)} & 87 $\pm$ 4 & 100 $\pm$ 0& \textbf{+14\%}\\
\midrule
\emph{alzheimer\_acetyl} & 66 $\pm$ 1 & 68 $\pm$ 1& \textbf{+3\%}\\
\emph{alzheimer\_amine} & 70 $\pm$ 1 & 76 $\pm$ 1& \textbf{+8\%}\\
\emph{alzheimer\_mem} & 64 $\pm$ 1 & 63 $\pm$ 1& -1\%\\
\emph{alzheimer\_toxic} & 72 $\pm$ 1 & 74 $\pm$ 1& \textbf{+2\%}\\
\midrule
\emph{wn18rr1} & 95 $\pm$ 1 & 98 $\pm$ 0& \textbf{+3\%}\\
\emph{wn18rr2} & 75 $\pm$ 1 & 79 $\pm$ 1& \textbf{+5\%}\\
\end{tabular}
\caption{Accuracy for \name{} with and without noisy constraints.}
\label{tab:constraints_acc_appendix}
\end{table}

\begin{table}[ht]
\centering
\footnotesize
\begin{tabular}{@{}l|ccc@{}}
\textbf{Task} & \textbf{Without} & \textbf{With} & \textbf{Difference}\\
\midrule
\emph{iggp-md (0)} & 14 $\pm$ 0 & 1 $\pm$ 0& \textbf{-92\%}\\
\emph{iggp-md (10)} & 109 $\pm$ 2 & 2 $\pm$ 0& \textbf{-98\%}\\
\emph{iggp-md (20)} & 103 $\pm$ 1 & 2 $\pm$ 0& \textbf{-98\%}\\
\emph{iggp-buttons (0)} & 61 $\pm$ 0 & 7 $\pm$ 0& \textbf{-88\%}\\
\emph{iggp-buttons (10)} & 61 $\pm$ 1 & 9 $\pm$ 0& \textbf{-85\%}\\
\emph{iggp-buttons (20)} & 57 $\pm$ 0 & 10 $\pm$ 0& \textbf{-82\%}\\
\emph{iggp-coins (0)} & 615 $\pm$ 5 & 138 $\pm$ 1& \textbf{-77\%}\\
\emph{iggp-coins (10)} & 631 $\pm$ 14 & 141 $\pm$ 2& \textbf{-77\%}\\
\emph{iggp-coins (20)} & 596 $\pm$ 2 & 144 $\pm$ 2& \textbf{-75\%}\\
\emph{iggp-rps (0)} & 195 $\pm$ 1 & 50 $\pm$ 1& \textbf{-74\%}\\
\emph{iggp-rps (10)} & 197 $\pm$ 1 & 60 $\pm$ 2& \textbf{-69\%}\\
\emph{iggp-rps (20)} & 193 $\pm$ 1 & 66 $\pm$ 1& \textbf{-65\%}\\
\midrule
\emph{zendo1 (0)} & 33 $\pm$ 9 & 13 $\pm$ 3& \textbf{-60\%}\\
\emph{zendo1 (10)} & 648 $\pm$ 3 & 77 $\pm$ 4& \textbf{-88\%}\\
\emph{zendo1 (20)} & 688 $\pm$ 7 & 100 $\pm$ 13& \textbf{-85\%}\\
\emph{zendo2 (0)} & 603 $\pm$ 4 & 48 $\pm$ 1& \textbf{-92\%}\\
\emph{zendo2 (10)} & 611 $\pm$ 1 & 48 $\pm$ 3& \textbf{-92\%}\\
\emph{zendo2 (20)} & 766 $\pm$ 70 & 118 $\pm$ 36 & \textbf{-84\%}\\
\emph{zendo3 (0)} & 613 $\pm$ 2 & 49 $\pm$ 3& \textbf{-92\%}\\
\emph{zendo3 (10)} & 626 $\pm$ 2 & 62 $\pm$ 2& \textbf{-90\%}\\
\emph{zendo3 (20)} & 834 $\pm$ 66 & 190 $\pm$ 112& \textbf{-77\%}\\
\emph{zendo4 (0)} & 594 $\pm$ 4 & 43 $\pm$ 3& \textbf{-92\%}\\
\emph{zendo4 (10)} & 616 $\pm$ 3 & 58 $\pm$ 2& \textbf{-90\%}\\
\emph{zendo4 (20)} & 767 $\pm$ 36 & 122 $\pm$ 32& \textbf{-84\%}\\
\midrule
\emph{dropk (0)} & 541 $\pm$ 5 & 7 $\pm$ 1& \textbf{-98\%}\\
\emph{dropk (10)} & \emph{timeout} & \emph{timeout}& 0\%\\
\emph{dropk (20)} & \emph{timeout} & \emph{timeout}& 0\%\\
\emph{evens (0)} & 770 $\pm$ 2 & 7 $\pm$ 0& \textbf{-99\%}\\
\emph{evens (10)} & \emph{timeout} & \emph{timeout}& 0\%\\
\emph{evens (20)} & \emph{timeout} & \emph{timeout}& 0\%\\
\emph{reverse (0)} & \emph{timeout} & 45 $\pm$ 7& \textbf{-96\%}\\
\emph{reverse (10)} & \emph{timeout} & \emph{timeout}& 0\%\\
\emph{reverse (20)} & \emph{timeout} & \emph{timeout}& 0\%\\
\emph{sorted (0)} & 1182 $\pm$ 8 & 31 $\pm$ 3& \textbf{-97\%}\\
\emph{sorted (10)} & \emph{timeout} & \emph{timeout}& 0\%\\
\emph{sorted (20)} & \emph{timeout} & \emph{timeout}& 0\%\\
\midrule
\emph{alzheimer\_acetyl} & \emph{timeout} & 133 $\pm$ 5& \textbf{-88\%}\\
\emph{alzheimer\_amine} & \emph{timeout} & 73 $\pm$ 3& \textbf{-93\%}\\
\emph{alzheimer\_mem} & \emph{timeout} & 79 $\pm$ 3& \textbf{-93\%}\\
\emph{alzheimer\_toxic} & \emph{timeout} & 61 $\pm$ 6& \textbf{-94\%}\\
\midrule
\emph{wn18rr1} & \emph{timeout} & 534 $\pm$ 14 & \textbf{-55\%}\\
\emph{wn18rr2} & \emph{timeout} & \emph{timeout}& 0\%\\

\end{tabular}
\caption{Learning time for \name{} with and without noisy constraints.}
\label{tab:constraints_time_appendix}
\end{table}

\end{appendices}
\end{document}